\pdfoutput=1
\PassOptionsToPackage{table,dvipsnames}{xcolor}
\documentclass[11pt]{article}
\usepackage{algorithmicx}

\usepackage[final]{acl}

\usepackage{times}
\usepackage{latexsym}

\usepackage[T1]{fontenc}
\usepackage[utf8]{inputenc}
\usepackage[inkscapeformat=pdf]{svg}

\usepackage{microtype}

\usepackage{inconsolata}

\usepackage{graphicx}

\usepackage{algorithm}
\usepackage{algpseudocode}

\usepackage{amsmath}
\usepackage{amssymb}
\usepackage{mathtools}
\usepackage{amsthm}
\usepackage[most]{tcolorbox}
\usepackage{graphicx}
\usepackage{booktabs}
\usepackage{caption}
\usepackage{makecell}

\usepackage[dvipsnames]{xcolor}
\usepackage{enumitem}

\usepackage{amsfonts}

\usepackage[most]{tcolorbox}
\usepackage{tabularx}
\usepackage{enumitem}
\usepackage{booktabs}

\newcolumntype{L}{>{\raggedright\arraybackslash}X}
\newcolumntype{C}{>{\centering\arraybackslash}X}

\newcommand{\name}{PILOT}

\usepackage{enumitem}
\usepackage{thmtools}
\usepackage{thm-restate}

\usepackage{booktabs}
\usepackage{colortbl}

\usepackage[capitalize,noabbrev]{cleveref}

\theoremstyle{plain}
\newtheorem{theorem}{Theorem}[section]
\newtheorem{proposition}[theorem]{Proposition}

\theoremstyle{definition}

\theoremstyle{remark}

\title{Adaptive LLM Routing under Budget Constraints}

\usepackage{authblk}
\makeatletter
\renewcommand\AB@authnote[1]{}   % remove author note marker
\renewcommand\AB@affilnote[1]{}  % remove affiliation note marker

\makeatother

\title{Adaptive LLM Routing Under Budget Constraints}

\title{Adaptive LLM Routing Under Budget Constraints}

\author[1]{\textbf{Pranoy Panda}\thanks{Equal contribution}}
\author[1]{\textbf{Raghav Magazine}\protect\footnotemark[1]\thanks{Current affiliation: Microsoft Research}}
\author[1]{\textbf{Chaitanya Devaguptapu}}
\author[1]{\textbf{Sho Takemori}}
\author[1]{\textbf{Vishal Sharma}\thanks{Current affiliation: Microsoft}}

\affil[1]{Fujitsu Research}

\begin{document}
\maketitle
\begin{abstract}
Large Language Models (LLMs) have revolutionized natural language processing, but their varying capabilities and costs pose challenges in practical applications. LLM routing addresses this by dynamically selecting the most suitable LLM for each query/task. Previous approaches treat this as a supervised learning problem, assuming complete knowledge of
optimal query-LLM pairings. However, real-world scenarios lack such comprehensive mappings and face evolving user queries.
We thus propose to study LLM routing as a contextual bandit problem, enabling adaptive decision-making using bandit feedback without requiring exhaustive inference across all LLMs for all queries (in contrast to supervised routing). To address this problem, we develop a shared embedding space for queries and LLMs, where query and LLM embeddings are aligned to reflect their affinity. This space is initially learned from offline human preference data and refined through online bandit feedback. 
We instantiate this idea through  \textit{\textbf{P}reference-prior \textbf{I}nformed \textbf{L}inUCB f\textbf{O}r Adaptive Rou\textbf{T}ing} (\name), a novel extension of LinUCB. 
To handle diverse user budgets for model routing, we introduce an online cost policy modeled as a multi-choice knapsack problem, ensuring resource-efficient routing. 
\end{abstract}

\section{Introduction}
\label{sec:intro}
Deploying Large Language Models (LLMs) in real-world systems faces a critical challenge: balancing performance with cost-effectiveness \cite{li2024llm}.  While larger models offer superior performance, their high costs makes their universal deployment impractical. This challenge is particularly acute given the varying pricing structures of proprietary models and the resource requirements of deploying the open-source alternatives.

\noindent To understand the need for varying resource requirements, consider a customer service chatbot handling diverse queries. For simple queries like ``\textit{What are your business hours?}", a smaller, cost-effective model might suffice. However, for complex inquiries, such as ``\textit{I'm torn between two of your smartphone models: the X200 and the Z300. I need a phone with excellent battery life, a high-quality camera, and robust performance for multitasking. Can you provide a detailed comparison, including any potential drawbacks of each model?}",  a more powerful (and costly) model may be necessary to ensure better planning and reasoning capabilities, which a smaller model might lack. This scenario illustrated the need for dynamic query routing -- The ability to dynamically route queries to the most appropriate model based on complexity and cost considerations, known as the \textit{model-routing} problem~\cite{ding2024hybrid}. 

\noindent Existing approaches model the routing problem as a supervised learning task, requiring large-scale labeled datasets mapping queries to their optimal LLM pairings ~\cite{ding2024hybrid, hu2024routerbench}. This problem formulation faces two limitations: (1) Gathering such labeled datasets is very expensive, as it requires responses from each model in the model pool for every query to discover the optimal query-LLM pairing. (2) Lack of adaptability to change in query distribution.

\noindent Thus, to find a more practical problem setting, we draw parallels with news recommendation systems, where models can only learn from user feedback, such as clicks on a single article \cite{li2010contextual, bouneffouf2019survey}. These systems must predict the best article for a user without showing all possible articles 
% \VS{update options to "possible articles"}
, receiving feedback solely on the selected article. Similarly, LLM routing involves choosing the best model for a query, where obtaining user feedback on all models is labor-intensive and costly. 

\noindent Building on this insight, \textit{we reformulate LLM routing as a contextual bandit learning problem} - a formulation commonly used in news and ad recommendation scenarios \cite{bouneffouf2019survey}. Thus, instead of requiring outputs from every LLM to identify the best match, our problem setting relies only on a binary bandit feedback, i.e., whether the chosen LLM's response is good or not. This approach is practical, as simple feedback mechanisms, like thumbs up/down ratings (i.e. like/dislike feedback), are now common in chat interfaces~\cite{appcues2024rating,delighted2024surveytypes}, allowing effective learning from user interactions without need for extensive annotation across LLMs.

\noindent To address this newly formulated problem of LLM routing with bandit feedback, we propose to develop an evolving shared embedding space for queries and LLMs, where distances represent routing affinity. Initially pretrained on human preference data \cite{chiang2024chatbot}, it is refined through online user feedback. Furthermore, to enforce cost constraints in an online setting, we introduce a novel policy modeled as an online multi-choice knapsack problem \cite{chakrabarty2008online}, dynamically allocating resources to balance budget adherence and performance.

\noindent Our key contributions are as follows,\\
(i) We formulate LLM routing as a budget constrained contextual bandit problem for adaptive decision-making with limited supervision \\ % (Section \ref{subsubsec:bandit}).\\
(ii) We propose a Preference-Prior Informed LinUCB algorithm (\name) that combines offline human preference data with online bandit feedback to route queries (Sec \ref{subsubsec:bandit}). We also show our preference prior helps achieve a lower regret bound than standard algorithm. This algorithm is further coupled with an online cost policy to dynamically allocate cost budget to queries (Sec \ref{subsubsec:cost_policy}).\\
(iii) In Sec \ref{sec:results-and-analysis} we show that our method outperforms existing bandit baselines across datasets, achieving lowest regrets and highest performance across different cost budgets.

\section{Methodology}
\setlength{\abovedisplayskip}{3pt}
\setlength{\belowdisplayskip}{3pt}

\vspace{-0.1cm}
\subsection{Problem Formulation}
% As discussed earlier, in this work, we address the problem of routing queries to large language models (LLMs) in an online setting with bandit feedback, aiming to maximize overall performance while adhering to user-defined budget constraints. We now provide the formal problem statement.

As discussed earlier, this work tackles the problem of routing queries to Large Language Models (LLMs) in an online setting, learning solely from evaluative (bandit) feedback from users. Leveraging these user interactions, our goal is to maximize overall performance under budget constraints, effectively personalizing LLM selection over time. We now present the formal problem statement.

\noindent Let $L = \{l_1, l_2, \ldots, l_k\}$ be a set of $k$ LLMs, $\mathcal{Q}$ be the space of all natural language queries, and $\mathcal{Y}$ be the space of all natural language responses. A query $q_t \in \mathcal{Q}$ at time $t$ is represented by its embedding $x_t \in \mathbb{R}^{d_{e}}$ in a $d_{e}$-dimensional space, generated by a pre-trained embedding model $\phi: \mathcal{Q} \rightarrow \mathbb{R}^{d_e}$. We assume black-box access to each LLM $l_i \in L$, where the response of LLM $l_i$ to query $q_t$ is denoted as $y_{t}^{l_i} \in \mathcal{Y}$. The quality of a response is quantified by a scoring function $s: \mathcal{Q} \times \mathcal{Y} \rightarrow [0,1]$, derived from human feedback or heuristic-based metrics.

\noindent At each time step $t=1, \dots, Q$, for a given query $q_t \in \mathcal{Q}$, an LLM router $M: \mathcal{Q}\rightarrow L$ selects an LLM $l \in L$. After receiving the response from the selected LLM $l$, the router observes a reward $r_t = s(q_t, y_t^{l}) \in [0,1]$ representing the quality of the response. Each LLM $l_i \in L$ also incurs a token cost $C_{t}^{l_i} \ge 0$ for processing query $q_t$. The objective of the LLM router is to maximize the total reward, defined as $\sum_{t=1}^Q r_t$, while satisfying the budget constraint $\sum_{t=1}^{Q} C_{t}^{M(q_t)} \le B$, i.e for \(Q\) consecutive queries, it ensures that the total token cost across these queries is less than \(B\).

\vspace{-2mm}
\subsection{Proposed Method}
\label{section:proposed_method}

We will now describe our solution approach to LLM routing that hinges on learning an effective mapping from queries to the LLMs.
For this, we will learn (i) an embedding of a given query and (ii) an embedding for each LLM in a shared embedding space such that the cosine distance between a query and an LLM represents their mutual affinity. The query-LLM shared embedding space is not static; rather, it evolves through online training.

\noindent While the online bandit feedback may be sufficient to train such an embedding space, it may still take a considerable amount of time to train in a completely online fashion. Further, a vast amount of public data is available in the form of human preferences where given a query and responses from two LLMs, humans provide their preferred LLM response. We hypothesize that independent of the end task (model routing in our case), this human preference data can be leveraged to pretrain the shared embedding space. The online bandit feedback can then be used at run time to continuously improve the pretrained embedding space, thus enhancing the accuracy of our routing decisions over time. Hence, we leverage two primary sources of information, (i) offline human preference data (Section \ref{subsubsec:human_pref}) to pretrain the shared embedding space and (ii) online bandit feedback (Section \ref{subsubsec:bandit}) to continuously tune the embedding space at runtime. Finally, to address the critical aspect of user-level budget constraints, we implement an online cost policy (Section \ref{subsubsec:cost_policy}). However, its worth noting that in the online bandit learning phase (Section \ref{subsubsec:bandit}), there is no budget constraint. - we elaborate on this in Section \ref{sec:exp_setup}. This multi-faceted approach allows us to balance performance optimization with budget adherence in a dynamic, user-centric manner. We now describe these steps in detail. For a birds eye view of the method, see \hyperref[algo:main]{Algorithm 1 \& 2}.

\begin{figure}[!h]
  \centering
    \includegraphics[width=1\columnwidth]{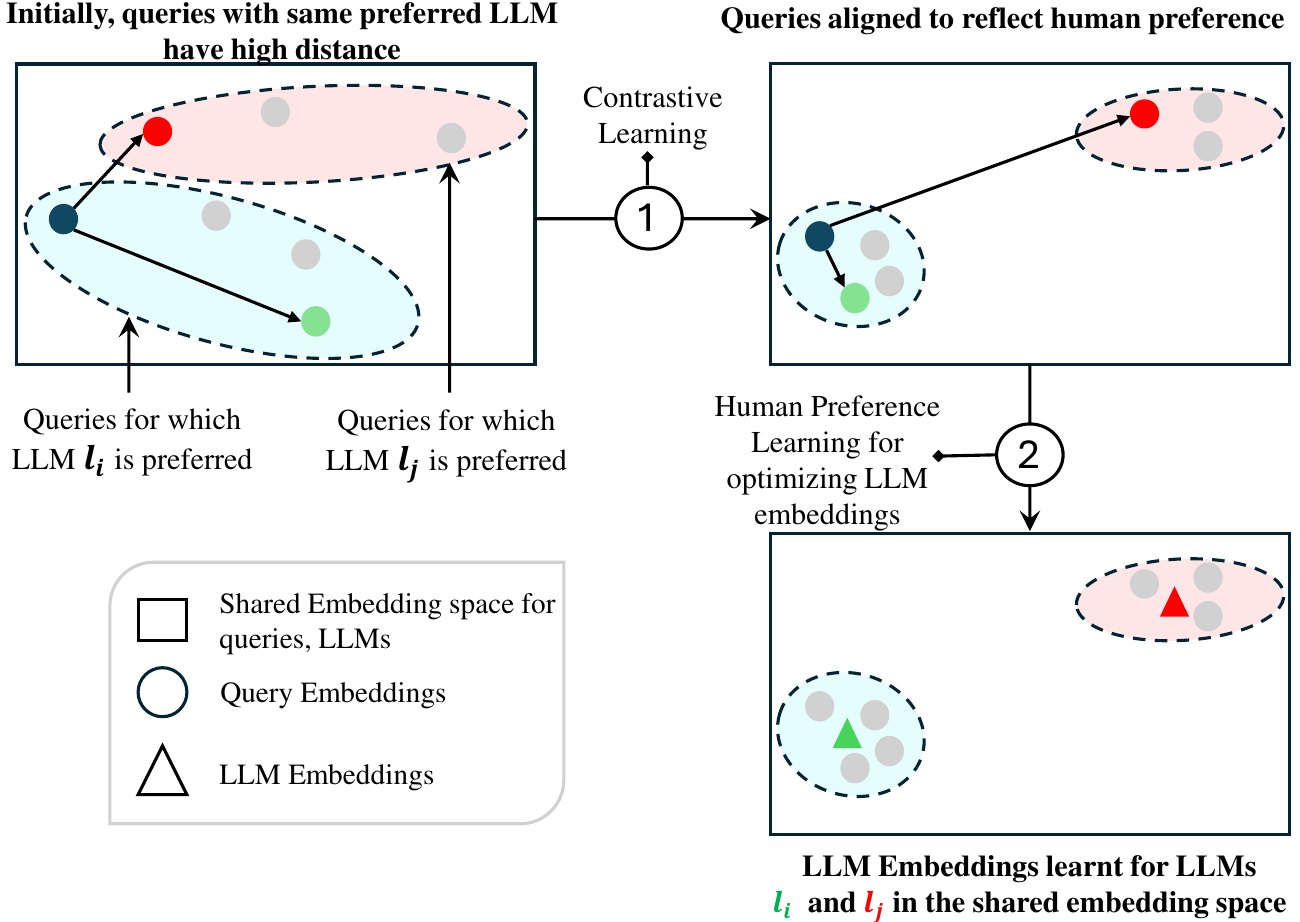}
\vspace{-0.4cm}  \caption{\textbf{Pretraining with Human Preference Data} {\Large{$\textcircled{\small{\textbf{1}}}$}} We leverage human preference dataset to learn query embeddings which are aligned w.r.t. human preferences on query-LLM mapping. Then, in {\Large{$\textcircled{\small{\textbf{2}}}$}} we learn LLM embeddings aligned with projected queries}
  \label{fig:method_offline}
  \vspace{-0.35cm}
\end{figure}

\subsubsection{Pretraining with Human Preferences}
\label{subsubsec:human_pref}
% Human preference data is a rich and diverse source of information for understanding query-LLM fit. We thus leverage it to provide a meaningful shared embedding space before incorporating online bandit feedback on the target task. The pretraining unfolds in two distinct phases. In phase one, we learn a projection from (initial) query embeddings onto our shared embedding space using contrastive learning with the goal to capture the relative preferences encoded in the human preference data. In phase two, using the same human preference data, we learn an embedding for each LLM that lies in the shared embedding space. We note that, while training the query embedding and LLM embeddings jointly is an option, in practice, we found that training in two separate phases leads to more stable training. We hypothesize that this is because training both the query projections and LLM embeddings jointly using the cosine similarity can lead to the moving target problem. 
Human preference data provides rich insights into query-LLM fit. We use it to establish a meaningful shared embedding space before incorporating online bandit feedback on the target task. Pretraining occurs in two phases for stability, mitigating the moving target problem that can arise from jointly optimizing query projections and LLM embeddings via cosine similarity. In phase one, we learn a projection from (initial) query embeddings onto our shared embedding space. In phase two, using same human preferences, we learn an embedding for each LLM that lies in the shared embedding space.

\noindent \textbf{Phase one: Learning the Query Projections} 

\noindent Given an existing query embedding model $\phi: \mathcal{Q} \rightarrow \mathbb{R}^{d_e}$, we project these embeddings to our $d_m$-dimensional shared space via a learned linear transformation $\psi(q) = W\phi(q) + b$. The parameters $W \in \mathbb{R}^{d_m \times d_e}$ and $b \in \mathbb{R}^{d_m}$ are learned using a cosine distance-based triplet loss on human preference data $D_{\text{pref}}$ (Figure \ref{fig:method_offline} {\Large{$\textcircled{\small{\textbf{1}}}$}}).
For each anchor query $(q_a, l_i, l_j, l_{win}) \in D_{\text{pref}}$ (where $l_{win}$ is the preferred LLM), we construct positive and negative query pools. The positive pool $P = \{(q, l_i, l_j, w) \in D_{\text{pref}} | \ \ l_w = l_{win}\}$ contains queries where $l_{win}$ was also preferred. The negative pool $N = \{(q, l_i, l_j, w) \in D_{\text{pref}} | \ \ l_w \neq l_{win} \land size(l_w)< size(l_{win})\}$ consists of queries where $l_{win}$ was not preferred against a smaller LLM (hard negatives, based on token cost - $\text{size}(l)$). 
% The triplet loss is:
% \vspace{-3mm}
% \begin{equation}
% \label{eqn:triplet_loss}
% \begin{split}
% L_{triplet} (q_a, P, N) = \max(0, \frac{1}{|N|}\sum_{q_n \in N}\cos(\psi(q_a), \psi(q_n))\\
% - \frac{1}{|P|}\sum_{q_p \in P} \cos(\psi(q_a), \psi(q_p)) + \epsilon )
% \end{split}
% \end{equation}\nobreak
% \noindent where $\epsilon$ is the margin.

\noindent \textbf{Phase two: Learning LLM Embeddings}
In phase two, we focus on learning the LLM embeddings $\theta_i$ for each LLM $l_i \in L$. For this, we first freeze the query projection parameters ($W$ and $b$) learned in phase one and then learn the LLM embeddings with the goal that given a query $(q, l_i, l_j, l_w) \in D_{\text{pref}}$, the embedding of the preferred LLM ($l_w$) is close to $\psi(q)$. We start by defining a probability distribution of $l_i$ winning over $l_j$ as $p_i = \frac{\exp(\cos(\theta_{i}, \psi(q)))}{\sum_{k \in \{i,j\}}\exp(\cos(\theta_{k}, \psi(q)))}$.
Then, we train the LLM embeddings by treating preference learning as a binary classification task using binary cross-entropy loss.
% as,
% \begin{equation}
% \label{eqn:bce_loss}
% L_{BCE} = -\sum_{(q, l_i, l_j, w) \in D_{\text{pref}}} 
% \Big[ \mathbb{I}_{w=i} \log(p_i) + \mathbb{I}_{w=j} \log(1 - p_i) \Big]
% \end{equation}
The final learned embeddings after this phase for an LLM $l_i$ is denoted as $\theta_i^{\text{pref}}$ (See {\Large{$\textcircled{\small{\textbf{2}}}$}} in Figure \ref{fig:method_offline}).

\noindent This two-phase learning process establishes an initial shared embedding space that captures the relationship between queries and LLMs based on human preferences, providing a strong foundation for subsequent online learning.

\begin{figure}[!h]
  \centering
    \includegraphics[width=1\columnwidth]{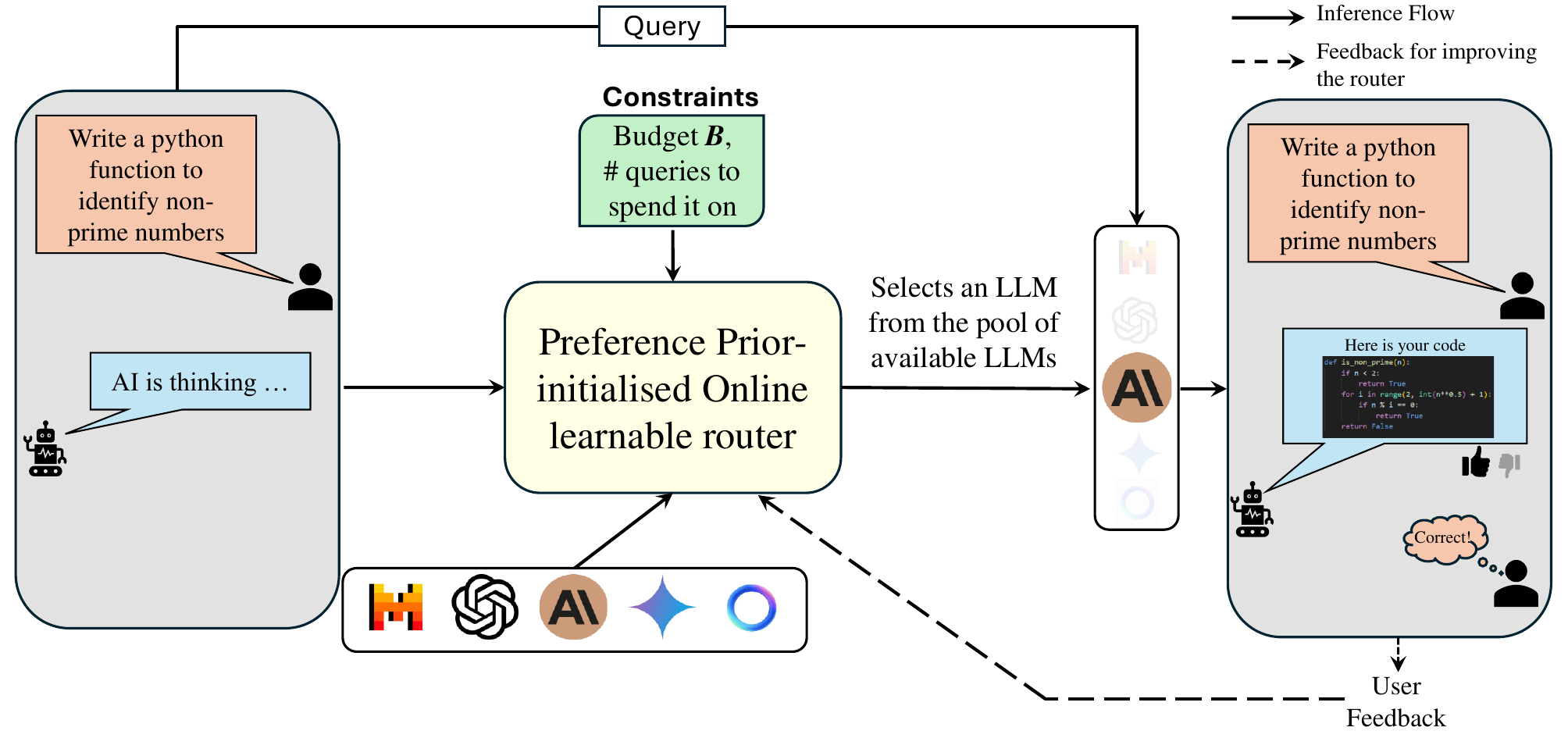}
    \vspace{-0.4cm}
  \caption{\textbf{Bandit Router Framework:} Our router takes three inputs: (i) User query (ii) cost constraints and, (iii) a model pool. It learns and adapts automatically based on user feedback, optimizing LLM selection over time.}
  \label{fig:method}
  \vspace{-0.35cm}
\end{figure}
% \vspace{-2mm}
\subsubsection{Evolving with Online Bandit Feedback}
\label{subsubsec:bandit}
Having learned query and LLM embeddings, we will now discuss how to incorporate the online bandit feedback for the end task of routing queries to appropriate LLMs. 
For this, we model routing as a contextual multi-armed bandit (CMAB) problem: projected query embeddings $\psi(q_t)$ serve as contexts, and LLMs are arms. Reward $r_t = s(q_t, y_t^{l}) \in [0,1]$ is the response quality from the selected LLM $l$. Objective is to maximize cumulative reward $\max_{\pi} \mathbb{E}\left[\sum_{t=1}^T r_t \right]$ via policy $\pi: \mathbb{R}^{d_m} \rightarrow \mathcal{A}$.

% we model the routing task as a contextual multi-armed bandit (CMAB) problem, where the projected query embedding at time $t$ (i.e. $\psi(q_t)$) serves as the context, and the available LLMs serve as the set of arms. We define reward $r_t = s(q_t, y_t^{l}) \in [0,1]$ representing the quality of the response of the selected LLM $l$. Our aim is to maximize cumulative reward over time:
% $\max_{\pi} \mathbb{E}\left[\sum_{t=1}^T r_t \right]$, where $\pi: \mathbb{R}^{d_m} \rightarrow \mathcal{A}$ is the router mapping contexts to arms.

\noindent The bandit's feedback is used to further align query and LLM embeddings with the goal of routing the query to appropriate LLM. For this, we choose to update arm/LLM embeddings at each round (following existing literature).

\noindent Let $\theta_a^t$ denote embedding of arm (LLM) $a$ at time step $t$ where the initial arm embedding at time $t=0$ is initialized with LLM embeddings learned in pretraining phase from preferences, i.e., $\theta^{0}_a = \theta_a^{\text{pref}}$.

\noindent As discussed in Section \ref{subsubsec:human_pref}, we model the mutual affinity between user queries and LLMs as the cosine distance between their representations in the shared embedding space.
In line with this idea, we model the expected reward at time $t$ in the CMAB problem as follows:

\vspace{-0.5cm}
\begin{center}
\begin{equation}
    \label{eqn:exp_reward}
    \mathbb{E}[r_t | a, q_t] = \cos(\hat{\psi}(q_t), \hat{\theta}_a) = \hat{\psi}(q_t) \cdot \hat{\theta}_a
\end{equation}
\end{center}
% \vspace{-0.5cm}
where $\hat{\psi}(q_t) = \frac{\psi(q_t)}{\|\psi(q_t)\|_2}$ and $\hat{\theta}_a = \frac{\theta_a}{\|\theta_a\|_2}$. Thus, owing to this linear reward formulation (cosine distance between unit normalized vectors), we propose a preference-prior informed linear upper confidence algorithm (\name) which builds upon the standard LinUCB~\cite{li2010contextual} algorithm while incorporating the knowledge gained from our preference learning phase.

\noindent \name, similar to LinUCB, performs online ridge regression. From a Bayesian perspective, this corresponds to maintaining a posterior distribution over the arm parameters. At each time step $t$, for each arm $a$, the point estimate of embedding of the arm (representing LLM $a$) is given by:
\begin{equation}
\begin{split}
\tilde{\theta}_a^t &= (A_a^t)^{-1}b_a^t 
\end{split}
\end{equation}
\noindent where, $A_a^t = A_a^{t-1} + \hat{\psi}(q_t)\hat{\psi}(q_t)^\top$, and, $b_a^t = b_a^{t-1} + r_t\hat{\psi}(q_t)$.  At $t=0$, we initialize the parameters $A_a^t$ and $b_a^t$ as, $A_a^0 = \lambda_a I, \quad b_a^0 = \lambda_a \theta_a^{\text{pref}}$, 
where $\theta_a^{\text{pref}}$ is the embedding for LLM corresponding to arm $a$ learned from preference data (see Section \ref{subsubsec:human_pref}), and $\lambda_a > 0$ is a regularization parameter.
This initialization can be interpreted as imposing a human preference prior $\hat{\theta}_a^0 \sim \mathcal{N}(\theta_a^{\text{pref}}, (\lambda_a I)^{-1})$ on the (initial) arm parameters. $\lambda_a$ controls prior strength: larger $\lambda_a$ implies less exploration (lower variance), while smaller $\lambda_a$ allows faster adaptation. To balance exploration, particularly if online queries differ from pretraining data, we set $\lambda_a$ as the inverse of arm $a$'s accuracy during the pretraining phase.

% The regularization parameter $\lambda_a$ controls the strength of this prior: a larger $\lambda_a$ induces less exploration for the arm, owing to the lower standard deviation of the arm parameter, while a smaller $\lambda_a$ allows for more rapid adaptation based on online feedback. However, the online queries can be quite different from the queries used for learning the initial arm/LLM embeddings. Thus, to ensure balanced exploration across all arms, $\forall a \in \mathcal{A}$ we set $\lambda_a$ to be inverse of each arm's training accuracy in the human preference learning phase

% (Section 2.2.1, Phase two).

\noindent With this formulation, the posterior distribution of arm (LLM) embeddings at time $t$ becomes: $p(\hat{\theta}_a^t | \mathcal{D}_t) = \mathcal{N}(\tilde{\theta}_a^t, (A_a^t)^{-1})$
, where $\mathcal{D}_t$ represents the observed online data up to time $t$.
The preference-prior informed LinUCB thus allows us to start with an informed estimate of LLM performance based on offline preference data, while still adapting to task dependent bandit feedback and query characteristics through online learning. 

The algorithm balances exploration and exploitation by selecting the arm that maximizes the upper confidence bound, i.e. $a_t$ is defined as:
\begin{equation*}
\arg \max_a \left(\cos(\hat{\psi}(q_t), \tilde{\theta}_a^t) + \alpha\sqrt{\hat{\psi}(q_t)^\top (A_a^t)^{-1} \hat{\psi}(q_t)}\right)
\end{equation*}
where $\alpha$ is the exploration parameter.

\newcommand{\thetapref}{\theta^{\mathrm{pref}}}
\newcommand{\cvec}[2]{x(#1, #2)}
\newcommand{\RR}{\mathbb{R}}
\newcommand{\trn}{\top}
\newcommand{\ofu}{\mathrm{OFUL}}
\newcommand{\ppiofu}{\mathrm{PI}\text{-}\mathrm{OFUL}}

\noindent To validate PILOT, we theoretically show that a preference-prior informed 
 bandit algorithm can achieve a smaller regret bound than the standard algorithm.
Here, we focus on Optimism in the face of uncertainty linear bandit algorithm (OFUL) ~\cite{abbasi2011improved} since both OFUL \& LinUCB are based on the same principle, i.e.,  principle of optimism in the face of uncertainty \citep[Chp 7.1]{lattimore2020bandit} \& OFUL is theoretically well-studied.
\begin{proposition}[Validity of a preference-prior informed bandit (informal)]
    Let $\thetapref \in \RR^{d'}$ and $\theta^* \in \RR^{d'}$ be 
    a pretrained vector and an unknown reward vector respectively.
    Let PI-OFUL be the OFUL \cite{abbasi2011improved} with the Preference-prior-Informed initialization with $\thetapref$.
    Then, if $\|\thetapref - \theta^*\| \le \|\theta^*\|$,
    PI-OFUL achieves a smaller cumulative regret bound than OFUL.
\end{proposition}
\vspace{-0.2cm}
\noindent See Sec \ref{subsec:appx_proof} for a more formal statement \& proof.

\vspace{-2.5mm}
%% NEEDS A LOT OF REFINEMENT

% Next, we discuss how to incorporate user budget constraints.

\subsubsection{Enforcing Budget Constraint with Online Cost Policy}
\label{subsubsec:cost_policy}

To manage user-specified cost budgets ($B$ over $Q$ queries), we introduce an online cost policy. This policy aims to optimally allocate the budget across unseen queries to maximize expected reward. We frame this as an online multi-choice knapsack problem (ON-MCKP) \cite{chakrabarty2008online}.

\noindent This ON-MCKP formulation allows leveraging the ZCL algorithm \cite{zhou2008budget} to enforce budget constraints while maximizing expected reward (Equation \ref{eqn:exp_reward}). In ON-MCKP, a knapsack of capacity $B$ receives item sets $N_t$ over time; at most one item (with value $v_j$ and weight $w_j$) is selected from each $N_t$ to maximize total value within $B$. \textit{In our context, at timestep $t$, available LLMs $L$ form the item set; their reward estimates $(\cos(\hat{\psi}(q_t), \hat{\theta}^{t}_l)$ $\forall l \in L)$ are values, and estimated token costs are weights}. We assume known upper/lower bounds ($UB, LB$) on the reward-to-cost ratio and query costs small relative to $B$, standard for online problems \cite{zhou2008budget}. The policy maintains budget utilization $z_t \in [0, 1]$. Following \cite{zhou2008budget}, eligible LLMs $E_t \subset L$ must satisfy $C^l_t \leq \frac{\cos(\hat{\psi}(q_t), \hat{\theta}^{t}_l)}{(\frac{UB \cdot e}{LB})^{z_t} (\frac{LB}{e})}$. We select the LLM with the highest expected reward from $E_t$ and update $z_t$.

\noindent Since the ZCL policy assumes an infinite horizon, potentially leading to underutilized budget over $Q$ queries, we implement a binning strategy. The $Q$ queries are partitioned into $N$ bins of size $S$, where $N = \lceil \frac{Q}{S} \rceil$ bin budget $= \frac{B}{N}$. The cost policy is applied per bin, with unused budget spilling over to the next, allowing flexible allocation within overall constraints. \hyperref[subsec:appx_algo_cost_policy]{Algorithm 3} (Appendix) details this policy, and \hyperref[algo:cost_policy]{Algorithm 2} provides an overview. \citet[Theorem 5.1]{zhou2008budget} establishes a performance guarantee, showing our online policy's performance is provably close to an optimal offline policy with full query knowledge.

% \begin{proposition}[\citet{zhou2008budget}] 
% The online cost policy ensures the budget constraint $\sum_{t=1}^{Q} C_{t}^{M(q_t)} \le B$.
% For any bin containing $S$ queries, the performance of our online cost policy, which uses the optimal ZCL algorithm, is within a factor of $\frac{1}{\ln (UB/LB) + 2}$ of the offline optimal performance.
% \end{proposition}
% \vspace{-0.2cm}

\begin{algorithm}[h!]
\caption{\name \ (\textbf{P}reference-prior \textbf{I}nformed \textbf{L}inUCB f\textbf{O}r Adaptive Rou\textbf{T}ing)}
\label{algo:main}
% \textbf{Input:} Human preference data $D_\text{pref}$, LLMs $L = \{l_1, l_2, \ldots, l_n\}$\\
\textbf{Input:} Human preference data $D_\text{pref}$, LLMs $L$\\
\vspace{0.04cm}
% Phase 1: Preference-Based Pretraining
\textit{\textbf{Preference-Based Pretraining}} \label{line:pref_based_pretraining}
\begin{algorithmic}[1]
\State Learn query projection $\phi$ by minimizing triplet loss using $(q_a, l_i, l_j, l_\text{win})$ tuples from $D_\text{pref}$ and constructing negative and positive samples as mentioned in Section \ref{section:proposed_method}.
\State Fix $\phi$ and learn LLM embeddings $\theta^{pref}_\text{LLM}$ using binary cross-entropy loss.
\vspace{0.02cm}

% Phase 2: Bandit Feedback-Based Online Fine-Tuning
\hspace*{-1.5cm}\textit{ \textbf{Online Bandit Learning}} \label{line:bandit_learning}
% \begin{algorithmic}[1]
\State Initialize bandit learning parameters $A_a = \lambda_a I$ and $b_a = \lambda_a \theta^{pref}_{a}$ for all $a \in L$.
\For{$t=1,\ldots, T$}
    \State Define $a_t$ as arm/LLM with largest UCB.
    \State Observe feedback $r_t$ w.r.t. response of selected LLM $a_t$ \& update parameters $A_a$ \& $b_a$.
\EndFor
\end{algorithmic}
\end{algorithm}

\vspace{-0.6cm} 

\begin{algorithm}[h!]
\caption{Online Cost Policy}
\textbf{Input:} budget $B$, query set length $Q$, LLMs $L$
\label{algo:cost_policy}
\begin{algorithmic}[1]
\For{$t=1,\ldots, Q$} % each query $q \in Q$  
    \State Estimate token cost $C_{t}^l \ \ \forall \ \ l \in L$ for $q_t$.  
    \State Compute the cost eligibility threshold for each LLM $l$:  
    $
    th_{t}^l = \frac{\cos(\hat{\psi}(q_t), \hat{\theta}^{t}_l)}{\left(\frac{UB \cdot e}{LB} \right)^{z_t} \cdot \left(\frac{LB}{e}\right)},
    $ 
    where $z_t \in [0,1]$ is the current budget utilization.  
    \State \textbf{yield} $l^{*} = \arg \max_{(l\in L \ \& \ C_{t}^l\leq th_{t}^l)} \cos(\hat{\psi}(q_t), \hat{\theta}^{t}_l)$
\EndFor
% \vspace{-1cm}
\end{algorithmic}
\end{algorithm}

\section{Experimental Setup}
\label{sec:exp_setup}
\vspace{-0.2cm}

\subsection{Evaluation Details}
To the best of our knowledge, we are the first to study LLM routing in an online bandit learning setting. Given the absence of an established experimental framework, we design the evaluation process by taking inspiration from \cite{li2010contextual}. Our objective is to simulate an online learning setting using an existing LLM routing dataset (Routerbench \cite{hu2024routerbench}).
We first split the routing dataset into \texttt{tuning} data (for hyperparameter selection) and \texttt{evaluation} data. The objective of \texttt{evaluation} data is to simulate online user query traffic. Similar to news recommendation \cite{li2010contextual}, when deploying the bandit routers to users, one reasonable way is to split all traffic into two buckets - (i) ``learning bucket”: a fraction of traffic on which various bandit algorithms are run to learn. (ii) The other, called ``deployment bucket”, is where we greedily serve users using bandit router obtained from learning bucket. 
% Below we explain the dataset details and the baselines.
% \vspace{-2mm}

\begin{table*}[!ht]
\renewcommand{\arraystretch}{1.2}
\setlength{\tabcolsep}{5pt}
\centering
\footnotesize
\begin{tabular}{>{\raggedright\arraybackslash}p{4.3cm} >{\raggedright\arraybackslash}p{9.5cm} >{\centering\arraybackslash}p{2.2cm}}
\toprule
\rowcolor{gray!20}
\multicolumn{1}{c}{\textbf{Experiment}} & \multicolumn{1}{c}{\hspace{-2cm} \textbf{Details}} & \multicolumn{1}{c}{\textbf{Reference}} \\
\midrule
\multicolumn{3}{c}{\cellcolor{blue!5}\textbf{\textsc{Main Experiments}}} \\
\textbf{Performance vs Budget Curves} &
Evaluates performance-cost trade-offs across budget constraints. &
\textit{Fig.~\ref{fig:single_domain_res} (a), (b)(i)} \\
\textbf{Performance with Varying Learning Data Sizes} &
Measures performance with different learning data quantities in deployment. &
\textit{Fig.~\ref{fig:single_domain_res} (a), (b)(ii)} \\
\textbf{Cumulative Regret Curves} &
Tracks learning efficiency over time compared to baselines. &
\textit{Fig.~\ref{fig:single_domain_res} (a), (b)(ii)} \\
\midrule
\multicolumn{3}{c}{\cellcolor{blue!5}\textbf{\textsc{Analysis}}} \\
\textbf{Qualitative Analysis of Routing} &
Examines routing decisions across diverse tasks (MMLU, MBPP, GSM8K). &
\textit{Section~\ref{subsec:routing_qual_analysis}} \\
\textbf{Compute Overhead of Routing} &
Measures routing latency overhead against GPT-4 inference time. &
\textit{Section~\ref{subsec:routing_compute_overhead}} \\
\textbf{Online Cost Policy Analysis} &
Compares adaptive online policy with fixed-budget and offline policies. &
\textit{Section~\ref{subsec:routing_cost_policy}} \\
\textbf{Embedding Model Sensitivity} &
Tests robustness across different query embedding models. &
\textit{Section~\ref{subsec:emb_model_sensitivity}} \\
\textbf{Comparison with Static Binary Supervised LLM Router} &
Contrasts with static supervised routers, especially under distribution shifts. &
\textit{Appendix~\ref{subsec:appx_comparison_with_hybrid_llm}} \\
\textbf{Human Preference Learning} &
Evaluates the human preference learning stage detailed in Section \ref{subsubsec:human_pref}. &
\textit{Appendix~\ref{appx:human_pref_analysis}} \\
\textbf{Ablation \& Sensitivity Analysis} &
Analyzes impact of individual components and robustness factors. &
\textit{Appendix~\ref{subsec:appx_addn_ablation_and_sens_anal}} \\
\bottomrule
\end{tabular}
\vspace{-0.2cm}
\caption{\textbf{Overview of Experiments:} Summary of empirical evaluations conducted to assess \name's performance.}
\label{tab:overview_experiments}
\vspace{-0.2cm}
\end{table*}

\vspace{-0.3cm}
\subsubsection{Dataset}
We evaluate our proposed method using Routerbench \cite{hu2024routerbench}, a comprehensive LLM routing dataset spanning a wide range of tasks including commonsense reasoning, knowledge-based language understanding, conversation, math, and coding. Routerbench is constructed by leveraging existing datasets (MMLU, Hellaswag, GSM8k, Winogrande, ARC Challenge, MTBench, MBPP) commonly used to evaluate leading LLMs. The dataset comprises $36,497$ samples covering $64$ tasks, with responses from $11$ different LLMs, including both open-source (Llama-70B-chat, Mixtral-8x7B-chat, Yi-34Bchat , Code Llama-34B, Mistral-7B-chat, WizardLM13B) and proprietary models (GPT-4, GPT-3.5-turbo, Claude-instant-v1, Claude-v1, Claude-v2). The dataset also includes the incurred cost and evaluation score (GPT-4 evaluation or exact match score, based on the task) of each LLM on each query. We request reader to see the original paper for details \cite{hu2024routerbench}. We use ChatArena \cite{chiang2024chatbot} for preference data, sampling a subset of queries where both associated LLMs belonged to RouterBench's set of $11$ LLMs.

\vspace{-0.2cm}
\subsubsection{Dataset Partitions} We partition the dataset as follows: we use $1000$ samples as \texttt{tuning} data for hyperparameter selection, with the remaining data split into ``learning" and ``deployment" buckets with a 10:1 ratio. This setup allows us to observe the router's performance improvement as more data becomes available in the learning bucket over time.
% \vspace{-0.3cm}
\subsubsection{Baselines}
\label{subsubsec:baselines}
% \vspace{-0.3cm}

We want to primarily understand two questions, 1) how well our approach performs against existing routers, and 2) in context of using a bandit setting, how well the use of \name \ 
 justifies against other choices of bandit algorithms.
% For the first question, we compare with \textit{all-to-one} routers where all the queries are fed to a single LLM. Additionally, in the Appendix \ref{subsec:appx_comparison_with_hybrid_llm}, we include a comparison with \textit{HybridLLM} \cite{ding2024hybrid}, a supervised binary LLM router, as a reference point, while keeping in mind the inherent differences in the problem formulation - for a given query, HybridLLM requires index of the optimal LLM as supervision, whereas we only assume evaluative feedback w.r.t. chosen LLM. For the second question, we compare against several contextual bandit baselines: \textit{LinUCB} \cite{li2010contextual} - contextual bandit algorithm that uses upper confidence bound with a linear reward formulation; \textit{Epoch-Greedy} \cite{langford2007epoch} - this algorithm alternates between exploration and exploitation phases; \textit{Explore Only}, a strategy that continuously explores without exploitation, an extreme case of epoch greedy; and a \textit{Random Policy} that selects LLMs randomly for every query.

% \noindent Its worth noting that to ensure fairness, we uniformly apply our proposed cost policy when comparing allocated budget against deployment set performance across different baselines (Figures \ref{fig:single_domain_res}, \ref{fig:embed_model_sensitivity}). We further access goodness of cost policy in Fig \ref{fig:cost_policy_eval}.

\begin{figure*}[ht]
  \centering
    \includegraphics[width=1\textwidth]{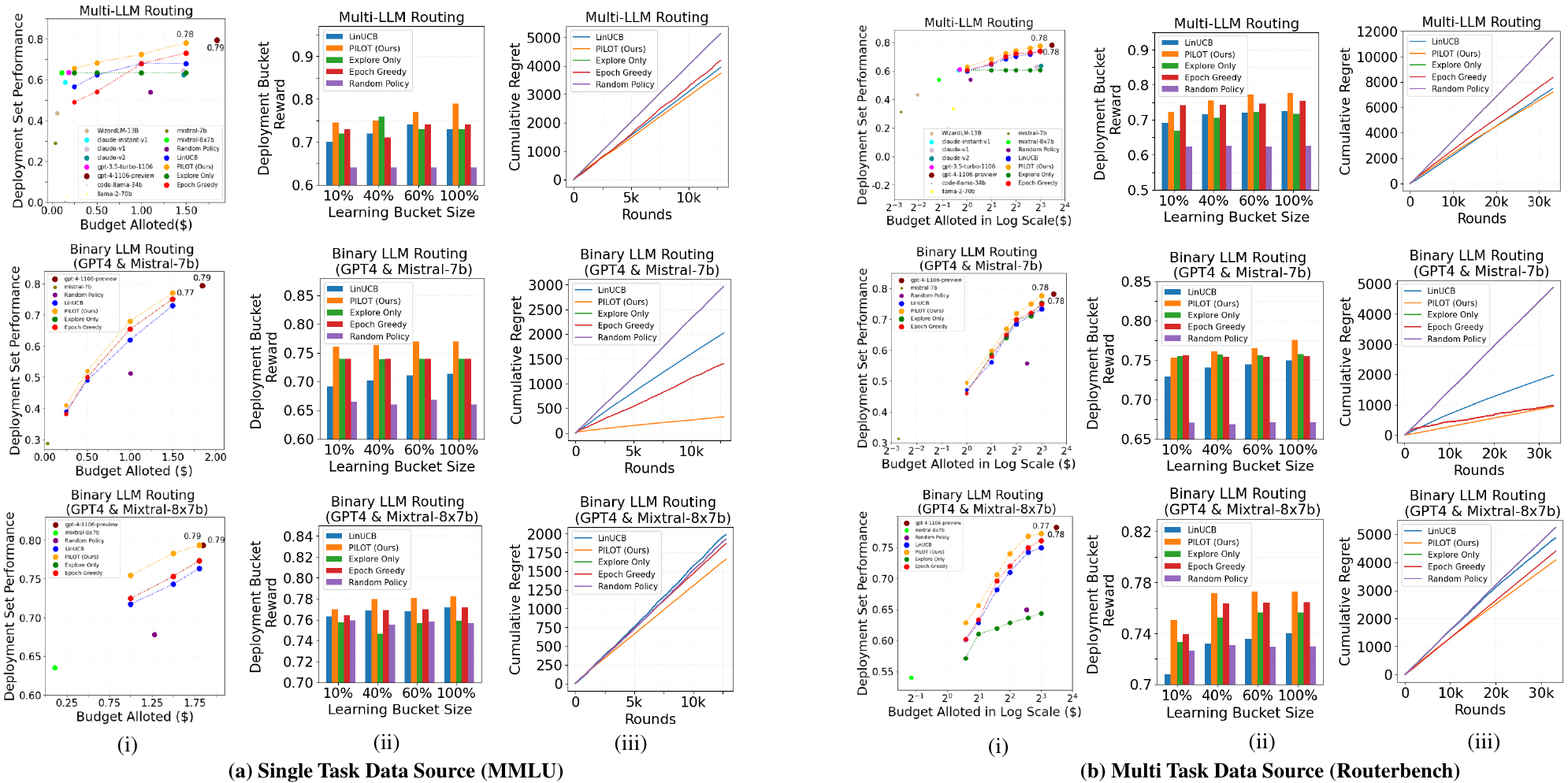}
    % \includesvg[width=1.0\textwidth]{images/main_results/main_result.svg}
    \vspace{-0.1cm}
  \caption{\textbf{Bandit Feedback based LLM Routing Evaluation}: In column (a) we report results for single task data source setting (MMLU), and in column (b) we report results for multi-task data source setting (Routerbench). The sub-column $(i)$ in each column represents performance vs cost curves on the held-out deployment set; sub-column $(ii)$ represents performance across different learning bucket sizes; sub-column $(iii)$ represents cumulative regret. Our method is shown in \textcolor{orange}{orange}.}
  \label{fig:single_domain_res}
  %\vspace{-0.35cm}
\end{figure*}

\begin{figure}[!ht]
  \centering
    \includegraphics[width=0.77\columnwidth]{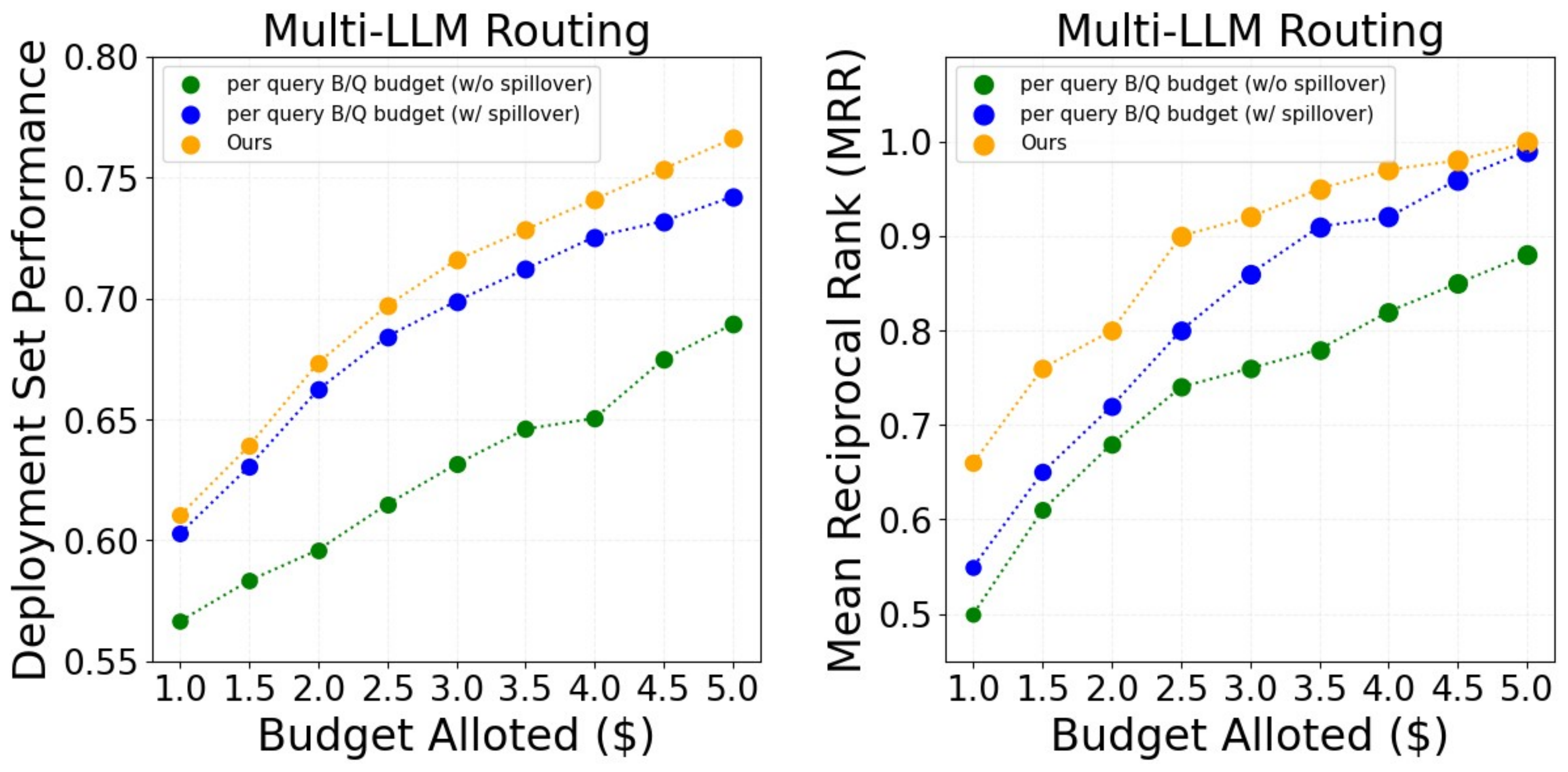}
    \vspace{-0.2cm}
  \caption{\textbf{Cost Policy Comparison}:
  % We compare our cost policy with baselines and show 
  (Left) Mean Reciprocal Rank of the chosen arms for various budgets  (Right) Performance of cost policies with diff budget}
  \label{fig:cost_policy_eval}
  \vspace{-0.4cm}
\end{figure}

\noindent For the first, we compare with \textit{all-to-one} routers (all queries to a single LLM). Appendix \ref{subsec:appx_comparison_with_hybrid_llm} also includes a reference comparison with \textit{HybridLLM} \cite{ding2024hybrid}, a supervised binary router, noting its different supervision requirement (optimal LLM index vs. our evaluative feedback).
For the second, we use several contextual bandit baselines: \textit{LinUCB} \cite{li2010contextual} (UCB with a linear reward model), \textit{Epoch-Greedy} \cite{langford2007epoch} (alternating exploration/exploitation), \textit{Explore Only} (continuous exploration), and a \textit{Random Policy} (random LLM selection).
Crucially, for fair comparison of allocated budget against deployment performance (Figures \ref{fig:single_domain_res}, \ref{fig:embed_model_sensitivity}), our proposed cost policy is uniformly applied across all baselines. The policy's effectiveness is further assessed in Fig \ref{fig:cost_policy_eval}.

\vspace{-0.25cm}
\subsection{Implementation Details}
\vspace{-0.1cm}
% embedding model - we use openai text-embedding-003-small for all experiments in Figure 2,3. To show the sensitivity of our method to embedding models we also conduction an additional exp in Sec ...
\noindent \textbf{Embedding Model}. We use OpenAI's \texttt{text-embedding-3-small} to embed queries for all results shown in Figure \ref{fig:single_domain_res}. To analyze the sensitivity of our router, we conduct an experiment using \texttt{Instructor-XL} (\citeauthor{su2023one}) in Section \ref{subsec:emb_model_sensitivity}.\\
% on the tuning data (1000 samples) what hyperparameters were chosen alpha -> grid search on (0.5, 1, 1.5, 2,5,10), reward coefficients -> grid search on (0.7, 0.8, 0.9), for estimating token cost we also compute the average LLM response token count and use that henceforth
\noindent \textbf{Hyperparameter selection}. 
 We use the tuning data to fine-tune hyperparameters of our method and baselines. 
% To find the optimal exploration hyperparameter $\alpha$ (for \name and LinUCB) we perform a grid search on values $\{1,1.5,2,5,10\}$ and choose the one that gives the maximum reward on the tuning data. For Epoch Greedy, we perform a grid search on $\{10,50,100,500\}$ to find optimal window size which we partition into exploration and exploitation steps. \\
To optimize the exploration parameter $\alpha$ (for \name \ and LinUCB), we search over $\{1, 1.5, 2, 5, 10\}$, selecting the value that maximizes reward. For Epoch Greedy, we do a grid search over window sizes $\{10, 50, 100, 500\}$ to find the optimal window size.

\noindent \textbf{Online Cost Policy}. During the deployment phase, our cost policy requires estimating the total cost of each query, including both input and output token costs. Input tokens are determined from the query itself, while output tokens are estimated using the mean output token count from responses in the tuning data for each LLM. This mean is then applied to all queries in the deployment set to calculate total query costs. Furthermore, It is worth noting that the proposed online cost policy operates independently of the \name\ algorithm. Its objective is to select the most suitable LLM for each query based on a query-wise LLM ranking, aiming to maximize the cumulative reward across \(Q\) queries while adhering to the total budget \(B\).

\vspace{-3mm}
\section{Results and Analysis}
\label{sec:results-and-analysis}

We evaluate \name's performance across various facets of LLM routing, considering diverse, multi-task applications and specific use cases. Our experiments utilize two data types:

\noindent (i) \textit{Single-task data source}: The MMLU benchmark from Routerbench, focusing on multi-choice question answering.
\noindent (ii) \textit{Multi-task data source}: Full Routerbench dataset, encompassing tasks like code generation, math problems, \& multi-turn conversations, simulating a broad range of user queries.

\noindent Using these datasets, we investigate several aspects of our bandit-based LLM routing algorithm, detailed under \textit{Main Experiments} in Table \ref{tab:overview_experiments}. Below, we summarize \name's performance for both single and multi-task scenarios, covering multi-LLM and binary-LLM routing settings. For binary-LLM routing, Figure~\ref{fig:single_domain_res} presents results for two LLM pairs, with more comparisons in Appendix~\ref{subsec:appx_addn_binary_llm_routing}.

% \vspace{-0.2cm}

\noindent \textbf{Multi-Task Data Source Setting}
Results in Figure \ref{fig:single_domain_res} column \textit{b}, indicate our method's superiority: We achieve a performance equal to $93\%$ of GPT-4's at just $25\%$ of its cost in multi-LLM setting while maintaining higher performance than any other baselines. Also, our method consistently shows the highest deployment set performance \& lower regret across various learning conditions.

\noindent \textbf{Single-Task Data Source Setting}
As shown in column \textit{a} of Figure \ref{fig:single_domain_res}, our method (\name) consistently outperforms all baseline bandit algorithms. In particular, in the multi-LLM routing case, we attain a performance of $86\%$ of GPT-4 at only $27\%$ of its cost and surpass all other all-to-one LLM baselines. \name \ also exhibits highest performance on deployment set across various learning bucket sizes, showing its efficacy with limited data. 
% The change in performance across learning bucket sizes is lower in binary LLM routing as less exploration is necessary owing to the two arms. W.r.t. the cumulative regret, our algorithm again demonstrates lower values in both multi \& binary LLM settings.

% \vspace{-4mm}

\vspace{-0.3cm}
\section{Discussion and Ablations}
% Here, we try to analyze, qualitatively and quantitatively, different aspects of the routing problem to give a holistic view of \name. For accessing these analysis easily we share them in Table \ref{tab:overview_experiments} under `Analysis'. \textcolor{blue}{TODO: add a one line summary of all the analysis before proceeding to next part}
This section provides qualitative and quantitative analyses of \name's routing behavior, computational efficiency, cost policy, and sensitivity, offering a holistic view. These analyses, summarized under `\textit{Analysis}' in Table \ref{tab:overview_experiments}, delve deeper into \name's operational characteristics.

\vspace{-0.2cm}
\subsection{Qualitative Analysis of PILOT's Routing}
\label{subsec:routing_qual_analysis}
Qualitative examination of \name's routing reveals intelligent decision-making. For demanding tasks like MMLU and ARC Challenge, \name \ routes $90\%$ and $89.4\%$ of queries to GPT-4, respectively, leveraging GPT-4's strength in complex reasoning. For coding tasks (MBPP), while GPT-4 is utilized, Claude models handle a significant $28\%$ of queries, indicating \name's recognition of Claude's coding abilities. In GSM8K, Claude (v1) is the predominant choice (94\% of queries). This isn't arbitrary; Claude-v1's strong performance on math tasks, coupled with its lower cost, makes it an effective option, echoing findings in \citeauthor{zhang2024careful}

\subsection{Analysis of the Online Cost Policy}
\label{subsec:routing_cost_policy}
\textbf{Comparison with other Online Cost Policies:}
Here we compare it to simple online baselines: (i) allocating $\frac{B}{Q}$ budget per query, and (ii) allocating $\frac{B}{Q}$ per query with spillovers from previous queries. We use these query-wise budgets to select the highest-ranked arm/LLM within the budget. We evaluate using mean reciprocal rank of the chosen arm (where rank 1 is the best arm w.r.t. learned bandit router) and performance on Routerbench's deployment set (multi-LLM case).

\begin{table}[h]
\centering
\small
\begin{tabular}{@{}c c c c@{}}
\toprule
\textbf{Cost (\$)} & $P - \lambda \mathcal{C}$ & \textbf{PILOT (Ours)} & \textbf{Difference} \\
\midrule
0.25 & 0.6079 & 0.6557 & \textbf{+0.0478} \\
0.50 & 0.6602 & 0.6840 & \textbf{+0.0238} \\
1.00 & 0.7265 & 0.7240 & -0.0025 \\
1.50 & 0.7740 & 0.7814 & \textbf{+0.0073} \\
\bottomrule
\end{tabular}
\caption{Comparison of performance between $P - \lambda \mathcal{C}$ offline policy and PILOT's online cost policy}
\label{tab:lamda_policy}
\vspace{-0.5cm}
\end{table}

% \vspace{-1cm}
\noindent \textbf{Comparison with Offline Cost Policy:} Here, we try to maximize $P-\lambda C$ (following \citealp{chen2022efficient}), where $P$ is estimated performance, $C$ is cost, \& hyperparameter $\lambda$ is tuned retrospectively for each budget constraint by optimizing its value over entire deployment set to achieve best possible outcome for that budget-a significant informational advantage. As shown in Table \ref{tab:lamda_policy}, our adaptive online policy generally outperforms this offline policy across multiple cost thresholds. This shows value of our online approach which performs comparably / better than even a policy with perfect hindsight. 

\vspace{-0.2cm}
\subsection{Computational Overhead of Routing}
\label{subsec:routing_compute_overhead}
\begin{table}[h]
\vspace{-0.2cm}
\centering
\small
\begin{tabular}{@{}p{2.5cm} p{2.3cm} p{1.9cm}@{}}
\toprule
\textbf{Embedding Model} & \textbf{Routing Time} & \textbf{GPT-4 Inference Time} \\
\midrule
\texttt{Instructor-XL} & 0.065 s & 2.5 s \\
% \midrule
OpenAI \texttt{text-} \texttt{embedding-3-small} & 0.239 s & 2.5 s \\
\bottomrule
\end{tabular}
\caption{\textbf{Analysis of Routing Time}: Routing Time refers to average time taken by \name \ to select a LLM from the pool and GPT-4 Inference Time is average time taken by GPT-4 to answer a query on MMLU dataset}
\label{tab:router-times}
\vspace{-0.3cm}
\end{table}

\noindent To assess \name's efficiency, we compare its average LLM selection time against GPT-4's average inference time on MMLU (Table \ref{tab:router-times}). \name's routing time is \textbf{10x} and \textbf{38x} faster than GPT-4 inference when using Instructor-XL \& OpenAI embeddings, respectively. This shows that \name adds negligible overhead to the response generation pipeline.

\vspace{-3mm}
\subsection{Embedding Model Sensitivity}
\label{subsec:emb_model_sensitivity}
\begin{figure}[H]
  \vspace{-0.3cm}
  \centering
    \includegraphics[width=0.75\columnwidth]{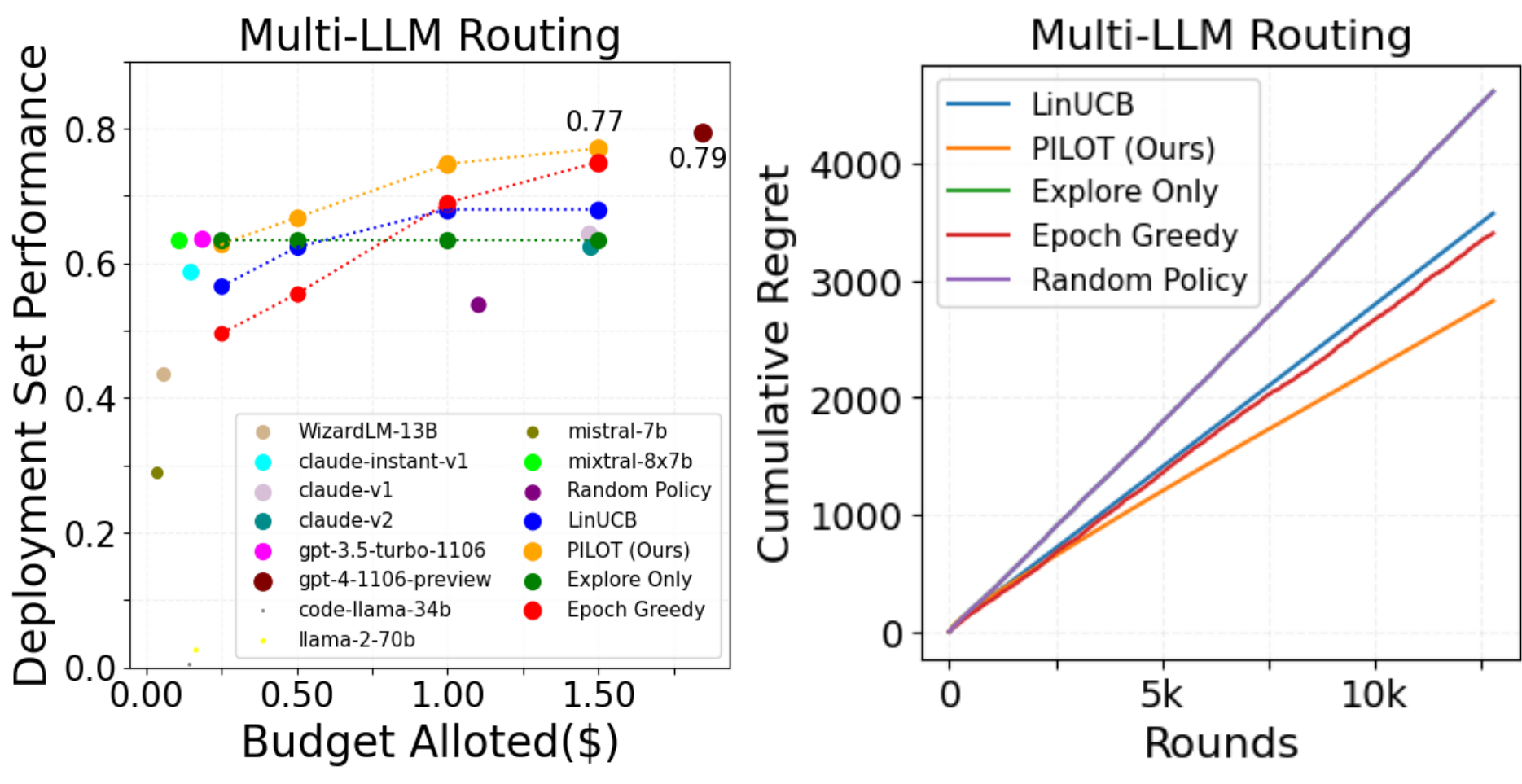}
    % \includesvg[width=0.75\columnwidth]{images/main_results/instructor_xl_sensitivity.svg}
  % \caption{\textbf{Embedding Sensitivity Analysis} \name \ performance comparison with bandit baselines using \texttt{Instructor-XL} embedding.}
  \caption{\textbf{Embedding Sensitivity Analysis} Figure compares \name's performance with bandit baselines using \texttt{Instructor-XL} embedding.}
  \label{fig:embed_model_sensitivity}
  \vspace{-0.6cm}
\end{figure}

% As stated in Section \ref{sec:exp_setup}, we used OpenAI's \texttt{text-} \texttt{embedding-3-small} model for query embeddings in all experiments.
% To assess \name's sensitivity to the query embedder used, in this experiment, we evaluate its performance on a different, albeit popular, open-source embedding model - \texttt{Instructor-XL} \cite{su2023one}. Results are shown in Figure \ref{fig:embed_model_sensitivity}. We notice that \name \ achieves superior performance w.r.t. baselines, indicating its robustness w.r.t. different embedding models.
\noindent Here, we assess \name's sensitivity to the embedder by evaluating its performance using another model - \texttt{Instructor-XL} \cite{su2023one}. As shown in Figure \ref{fig:embed_model_sensitivity}, \name \ continues to maintain superior performance over baselines.

\vspace{-0.1cm}
\section{Related Works}
\label{sec:rel_works}
\vspace{-0.2cm}
% Prior work on LLM routing and efficient ML (neighboring topic to routing), span three main areas: (1) Traditional efficiency methods like pruning \cite{lecun1989optimal}, quantization \cite{jacob2018quantization}, LoRA \cite{hu2021lora}, and distillation \cite{hinton2015distilling} offer static optimizations unsuited for varying constraints. (2) Hybrid approaches include routing methods \cite{ding2024hybrid} that assume larger models are always better, and synthesis methods \cite{jiang2023llm, lu2024blending} that combine responses at higher cost. (3) Existing routing strategies are either non-predictive, like FrugalGPT \cite{chen2023frugalgpt} which queries models sequentially until meeting quality thresholds, or predictive \cite{shnitzer2023large, narayanan2023tryage} requiring full supervision. In contrast, we treat routing as a contextual multi-armed bandit problem with budget constraints, learning from bandit feedback to handle dynamic query distributions without assuming model size determines performance. See Appendix \ref{sec:appx_related_works} for a more detailed survey.
Research on efficient LLM deployment spans static model optimizations, hybrid strategies, and dynamic routing. (1) \textit{Static efficiency methods}, such as pruning~\cite{lecun1989optimal}, quantization~\cite{jacob2018quantization}, LoRA~\cite{hu2021lora}, and distillation~\cite{hinton2015distilling}, compress models for fixed cost constraints, but cannot adapt to varying task demands. (2) \textit{Hybrid approaches} like LLM-Blender~\cite{lu2024blending} synthesize outputs from multiple LLMs to improve quality, while others such as TensorOpera Router~\cite{stripelis2024tensoropera} and FORC~\cite{vsakota2024fly} train offline meta-models to predict the best model. These require full supervision and may not generalize to new query distributions. (3) \textit{Routing strategies} select a single model per query. Non-predictive methods like FrugalGPT~\cite{chen2023frugalgpt} use sequential evaluation, while predictive ones like HybridLLM~\cite{ding2024hybrid}, GraphRouter~\cite{feng2024graphrouter}, and Confidence Tokens~\cite{chuang2025learning} train supervised routers with full feedback or LLM modifications.\\
In contrast, our work formulates routing as a contextual bandit problem with budget constraints. We learn an embedding-based router that adapts online using bandit feedback—observing reward only for the selected model—without requiring exhaustive supervision or query-specific full inference. This enables efficient and adaptive LLM deployment in dynamic, cost-sensitive environments.

\vspace{-0.1cm}
\section{Conclusion}
We address LLM routing with budget constraints using bandit feedback in this work. We propose \name, a human-preference prior based contextual bandit algorithm, coupled with a novel online cost policy that optimizes budget allocation across queries. Our approach achieves 93\% of GPT-4's performance at 25\% of its cost on Routerbench. 

\section*{Limitations}
During the online bandit learning (Algorithm \ref{algo:main}) we do not consider budget constraint, rather only during deployment we consider budget constraint. Underlying rational for decoupling the bandit algorithm and the cost policy was to ensure deployment stability and provide direct, user-controllable budget management. This separation facilitates deploying a robust bandit model while dynamically adjusting cost policy in real-time based on budget. However, one maybe be interested in learning under budget constraints, which we leave for future work.

\noindent In this work we focused on single-turn conversations as input for routing, however real-world scenarios could multi-turn interaction based routing. We leave this for future work.

% \section*{Acknowledgments}

% This document has been adapted
% by Steven Bethard, Ryan Cotterell and Rui Yan
% from the instructions for earlier ACL and NAACL proceedings, including those for
% ACL 2019 by Douwe Kiela and Ivan Vuli\'{c},
% NAACL 2019 by Stephanie Lukin and Alla Roskovskaya,
% ACL 2018 by Shay Cohen, Kevin Gimpel, and Wei Lu,
% NAACL 2018 by Margaret Mitchell and Stephanie Lukin,
% Bib\TeX{} suggestions for (NA)ACL 2017/2018 from Jason Eisner,
% ACL 2017 by Dan Gildea and Min-Yen Kan,
% NAACL 2017 by Margaret Mitchell,
% ACL 2012 by Maggie Li and Michael White,
% ACL 2010 by Jing-Shin Chang and Philipp Koehn,
% ACL 2008 by Johanna D. Moore, Simone Teufel, James Allan, and Sadaoki Furui,
% ACL 2005 by Hwee Tou Ng and Kemal Oflazer,
% ACL 2002 by Eugene Charniak and Dekang Lin,
% and earlier ACL and EACL formats written by several people, including
% John Chen, Henry S. Thompson and Donald Walker.
% Additional elements were taken from the formatting instructions of the \emph{International Joint Conference on Artificial Intelligence} and the \emph{Conference on Computer Vision and Pattern Recognition}.

% Bibliography entries for the entire Anthology, followed by custom entries
%\bibliography{anthology,custom}
% Custom bibliography entries only
\bibliography{custom}

\appendix
\clearpage
\section{Appendix}
\label{sec:appendix}

In this section, we provide additional results and details that we could not include in the main paper due to space constraints.
In particular, this appendix contains the following:

\vspace{0.3cm}
\begin{itemize}[nosep, leftmargin=*, label={}]
    \item {\large\textbf{\hyperref[sec:appx_related_works]{Extended Related Works}}}
    \begin{itemize}[nosep, leftmargin=1.5em]
        \item {\normalsize\hyperref[subsec:efficient_llm]{Efficient LLM Inference}}
        \item {\normalsize\hyperref[subsec:hybrid_llm]{Hybrid LLM Approaches}}
        \item {\normalsize\hyperref[subsec:routing_strategies]{Routing Strategies for LLMs}}
    \end{itemize}

    \vspace{1em}
    \item {\large\textbf{\hyperref[sec:appx_theoretical]{Theoretical Analysis and Algorithms}}}
    \begin{itemize}[nosep, leftmargin=1.5em]
        \item {\normalsize\hyperref[subsec:appx_proof]{Theoretical Analysis of a Preference-Prior-Informed Bandit Algorithm}}
        \item {\normalsize\hyperref[subsec:appx_algo_cost_policy]{Algorithm for the Online Cost Policy}}
    \end{itemize}

    \vspace{1em}
    \item {\large\textbf{\hyperref[sec:appx_experiments]{Additional Results and Analysis}}}
    \begin{itemize}[nosep, leftmargin=1.5em]
        \item {\normalsize\hyperref[subsec:appx_comparison_with_hybrid_llm]{Comparison with Supervised Binary Router}}
        \begin{itemize}[nosep, leftmargin=1.5em]
            \item {\small\hyperref[subsec:appx_comparison_with_hybrid_llm_perf_vs_cost]{Performance vs Cost Analysis}}
            \item {\small\hyperref[subsec:appx_adapt_to_shift_in_query_dist]{Adaptability to Shift in Query Distribution}}
        \end{itemize}
        \item {\normalsize\hyperref[appx:query_complexity_analysis]{Query Complexity Analysis}}
        \item {\normalsize\hyperref[appx:human_pref_analysis]{Analysis of Human Preference Learning}}
        \item {\normalsize\hyperref[subsec:appx_addn_binary_llm_routing]{Additional Binary LLM Routing Results}}
        \item {\normalsize\hyperref[subsec:appx_addn_ablation_and_sens_anal]{Ablation and Sensitivity Analysis}}
    \end{itemize}
\end{itemize}

\section{Extended Related Works}\label{sec:appx_related_works}
In Section \ref{sec:rel_works} of the main paper we briefly described works in the areas of routing strategies for language models and neighbouring research topics such as efficient LLM inference and hybrid LLM approaches. Here we elaborate on it.
\vspace{-2mm}
\subsection{Efficient LLM Inference}\label{subsec:efficient_llm}
Traditional approaches to efficient ML inference can be categorized into: model pruning \cite{lecun1989optimal}, quantization \cite{jacob2018quantization}, linear attention \cite{arorasimple}, low-rank adaptation \cite{hu2021lora} and knowledge distillation \cite{hinton2015distilling}. These methods typically produce static optimizations, which may not suffice for LLMs serving a range of tasks with varying accuracy/cost constraints. Our work, in contrast, focuses on dynamic optimizations to meet diverse user demands.
\vspace{-2mm}
\subsection{Hybrid LLM Approaches}\label{subsec:hybrid_llm}
% Recent work has introduced hybrid inference paradigms leveraging both small and large models. These can be classified into:
% \noindent (i) \textit{LLM routing}: In this approach one LLM is chosen from the available pool, to resolve the query. The Hybrid LLM approach \cite{ding2024hybrid} introduces a mechanism to route between a small and a large LLM, assuming the large LLM always outperforms the small one. (ii) 
% \textit{Multi-LLM synthesis}: This approach involves querying multiple LLMs and then combining their response to get the final response. Methods like \cite{jiang2023llm} and \cite{lu2024blending} utilize multiple LLMs to generate and synthesize responses, often incurring higher costs and latency due to the two step process. Our work extends these ideas by considering multiple LLMs without the fixed performance assumptions (i.e. a larger LLM is better than a smaller one for all queries) and aims to maintain high performance while calling only a single LLM per query.
Hybrid inference methods attempt to balance cost and quality by combining outputs or using meta-models. These include:

\noindent (i) \textit{Multi-LLM Synthesis}: LLM-Blender~\cite{lu2024blending} and related methods~\cite{jiang2023llm} invoke several models and fuse their responses. While improving output quality, these approaches are cost-intensive and unsuitable for latency-sensitive applications.

\noindent (ii) \textit{Meta-routing via reward estimation}: TensorOpera Router~\cite{stripelis2024tensoropera} builds a separate reward model to guide routing decisions over multiple LLMs. However, it relies on offline data and full supervision to train the reward predictor.

\noindent Unlike these works, we focus on selecting a single model per query and improve performance through online learning from partial feedback.
\vspace{-2mm}
\subsection{Routing Strategies for LLMs}\label{subsec:routing_strategies}
Existing routing strategies can be categorized as: 

\noindent (i) \textit{Non-predictive routing}: FrugalGPT~\cite{chen2023frugalgpt} executes models sequentially until a quality threshold is met. While simple, this leads to multiple model calls and doesn't generalize across queries.

\noindent (ii) \textit{Predictive routing (supervised)}: These methods train a router to choose among LLMs based on full supervision. HybridLLM~\cite{ding2024hybrid} trains a classifier to select between a small and large model. FORC~\cite{vsakota2024fly} uses a meta-model to balance accuracy and cost. GraphRouter~\cite{feng2024graphrouter} encodes queries and models into a bipartite graph to guide routing. Chuang et al.~\cite{chuang2025learning} introduce confidence tokens emitted by LLMs to help with model selection, requiring LLM modification.

\noindent While these methods show strong performance, they rely on full model evaluation during training, limiting scalability and adaptability. Our work differs by treating routing as a contextual bandit problem: we learn only from the selected model's feedback, adapting online to shifting distributions without exhaustive supervision.

\noindent Below we summarize our unique aspects to contextualize our work within LLM routing literature:
\begin{itemize}[nosep]
\item \textbf{Online Adaptation}: We train the router online, enabling it to adjust to evolving query types and workloads.
\item \textbf{Bandit Feedback}: We operate under partial supervision, learning from the reward of only the selected model, unlike prior work requiring all-model inference for each training query.
\item \textbf{Budget-Aware Routing}: Our formulation includes a cost policy via a multi-choice knapsack, explicitly managing user budgets.
\end{itemize}
This makes our method suitable for practical LLM deployment settings that demand efficiency, adaptability, and minimal supervision.

\section{Theoretical Analysis and Algorithms}\label{sec:appx_theoretical}

\subsection{Theoretical Analysis of a Preference-Prior-Informed Bandit Algorithm}
\label{subsec:appx_proof}
To show the validity of \name, we theoretically show that a preference-prior-informed 
 bandit algorithm can achieve a smaller regret bound than the standard algorithm.
Here, we focus on preference-prior informed OFUL \cite{abbasi2011online} since OFUL is a theoretically well-studied method, and both the  algorithms (LinUCB and OFUL) share the same principle, i.e., the principle of optimism in the face of uncertainty \citep[Chp 7.1]{lattimore2020bandit} 
and \cite{li2010contextual} have not provided a theoretical analysis of LinUCB.
We briefly introduce a problem setting for theoretical analysis.
For $t=1, \dots, T$, a query $q_t$ and selected LLM $l_t$, we assume the following reward model:
\begin{equation*}
    r_t(l_t, q_t) =  \theta^* \cdot \cvec{l_t}{q_t} + \varepsilon_t,
\end{equation*}
where $\cvec{l_t}{q_t} \in \RR^{d'}$ is a context vector and $\theta^* \in \RR^{d'}$ is an unknown reward vector.
We define cumulative regret 
\begin{equation*}
    R(T) = \sum_{t=1}^{T} \left( \max_{l \in L} r_t(l, q_t) - r_t(l_t, q_t) \right).
\end{equation*}
For $\lambda > 0$, 
we define an estimation $\hat{\theta}_t$ of $\theta^*$ as 
$A_t^{-1} \sum_{s=1}^{t-1}r_s \cvec{l_s}{q_s}$,
where $A_t = \lambda I + \sum_{s=1}^{t-1} \cvec{l_s}{q_s}\cvec{l_s}{q_s}^\trn$.
For $\delta \in (0, 1)$ and $S > 0$, we define a confidence set $\mathcal{C}_t(\delta, \hat{\theta}_t; S)$ by
% \begin{equation*}
%     \mathcal{C}_t(\delta, \hat{\theta}_t; S) = \{\theta \in \RR^{d'}: (\theta - \hat{\theta}_t)^\trn A_t (\theta - \hat{\theta}_t) \le 
%     \sqrt{\lambda}S
%         + R \sqrt{2 \log (1/\delta) + d' \log (1 + T/(\lambda d'))}\}.
% \end{equation*}
\begin{align*}
    \mathcal{C}_t(\delta, \hat{\theta}_t; S)
    = \Big\{ \theta \in \RR^{d'} : 
    \; (\theta - \hat{\theta}_t)^\trn A_t (\theta - \hat{\theta}_t) \le \\
    \sqrt{\lambda} S 
    + R \sqrt{2 \log (1/\delta) + d' \log \left(1 + \frac{T}{\lambda d'}\right)}
    \Big\}.
\end{align*}

Then for each round $t$, OFUL selects $l_t \in L$
such that $\max_{\theta \in \mathcal{C}_t(\delta, \hat{\theta}_t; S)}\theta \cdot x(l_t, q_t) = \max_{(\theta, l) \in \mathcal{C}_t(\delta, \hat{\theta}_t; S) \times  L} \theta \cdot x(l, q_t)$.

Then, if $\|\theta^*\| \le S$, OFUL has the following regret bound with probability at least $1 - \delta$,
\begin{equation*}
    R(T) \le U_T(S)
\end{equation*}
where 
% \begin{equation*}
% U_T(S) = 4 \sqrt {T d' \log (\lambda + T/d')} \left(
%         \sqrt{\lambda}S
%         + R \sqrt{2 \log (1/\delta) + d' \log (1 + T/(\lambda d'))}
%         \right).
% \end{equation*}
\begin{align*}
U_T(S) =\; 4 \sqrt{T d' \log (\lambda + T/d')} \cdot \Bigg(
    \sqrt{\lambda} S \\
 + R \sqrt{2 \log (1/\delta) + d' \log \left(1 + \frac{T}{\lambda d'} \right)}
\Bigg)
\end{align*}
If we know the optimal choice of $S$ (i.e., $S = \|\theta^*\|$), then regret bound $U_{T}(\ofu)$ of OFUL is given as 
\begin{equation*}
    U_{T}(\ofu) := U_{T}(\|\theta^*\|).
\end{equation*}

\begin{proposition}
 Let $\thetapref \in \RR^{d'}$ be a pretrained vector.
    We define an estimation $\tilde{\theta}_t $ of $\theta^*$
    with an initialization $\thetapref$
    as 
    $\tilde{\theta}_t := (A_{t})^{-1}b_t$,
where $A_t = \lambda I + \sum_{s=1}^{t-1} \cvec{l_s}{q_s}\cvec{l_s}{q_s}^\trn$, 
$b_t = \lambda \thetapref + \sum_{s=1}^{t-1} \cvec{l_s}{q_s}$.
We define PI-OFUL as OFUL with the confidence set $\mathcal{C}_t(\delta, \tilde{\theta}_t; S')$ with a parameter $S' > 0$, that is, PI-OFUL selects $l_t \in L$ such that 
$\max_{\theta \in \mathcal{C}_t(\delta, \tilde{\theta}_t; S')}\theta \cdot x(l_t, q_t) = \max_{(\theta, l) \in \mathcal{C}_t(\delta, \tilde{\theta}_t; S')) \times  L} \theta \cdot x(l, q_t)$.
    If $\| \theta^* - \thetapref\| \le S'$, then regret bound of PI-OFUL is given as 
    $U_T(S')$.
    In particular, if we know the optimal choice of $S'$ (i.e., $S' = \| \theta^* - \thetapref\|$), then
    then regret bound $U_{T}(\ppiofu)$ of PI-OFUL is given as 
    \begin{equation*}
        U_{T}(\ppiofu) := U_T(\| \theta^* - \thetapref\| ).
    \end{equation*}
    Thus, if $\| \theta^* - \thetapref\|  \le \|\theta^*\|$, we have
     \begin{equation*}
        U_{T}(\ppiofu) \le U_{T}(\ofu).
    \end{equation*}
\end{proposition}
\begin{proof}
    For $1 \le t \le T$, let $X \in \RR^{(t-1) \times d'}$ be the matrix whose rows are given as $\cvec{a_1}{q_1}^\trn, \dots, \cvec{a_{t-1}}{q_{t-1}}^\trn$.
    By definition of $\tilde{\theta}_t$ and the proof of \citep[Theorem 8]{abbasi2011online} and,
    $\tilde{\theta}_t$ is given as 
    \begin{equation*}
        \tilde{\theta}_t 
        = (X^\trn X + \lambda I)^{-1} X^\trn \varepsilon + \theta^* - \lambda (X^\trn X + \lambda I)(\theta^* - \thetapref),
    \end{equation*}
    where $\varepsilon \in \RR^{t-1}$ is defined as $\varepsilon^\trn = (\varepsilon_1, \dots, \varepsilon_{t-1})$.
    Thus, by the proof of \citep[Theorem 8]{abbasi2011online},
    we have the following with probability $1-\delta$
    % \begin{equation*}
    %    |\tilde{\theta}_t \cdot x - \theta^*\cdot x| \le \|x \|_{A_t^{-1}}
    %    \left(
    %    R \sqrt{2 \log\left(
    %    \frac{\det A_t^{1/2}\det(\lambda I)^{-1/2}}{\delta}
    %    \right)}
    %    + \lambda^{1/2}\left\|\theta^* - \thetapref\right\|
    %    \right).
    % \end{equation*}
    \begin{align*}
    &|\tilde{\theta}_t \cdot x - \theta^* \cdot x|
    \le \\
    &\quad \|x\|_{A_t^{-1}} \Bigg(
        R \sqrt{2 \log\left(
            \frac{\det(A_t)^{1/2} \det(\lambda I)^{-1/2}}{\delta}
        \right)} \\
    &\quad\quad+ \sqrt{\lambda} \left\| \theta^* - \thetapref \right\|
    \Bigg).
\end{align*}
    Using this confidence bound, by the standard argument (the same proof as \citep[Theorem 13]{abbasi2011online}),
    we have our assertion.
\end{proof}

\subsection{Algorithm for the Online Cost Policy}\label{subsec:appx_algo_cost_policy}

% \vspace{-0.1cm}
\begin{algorithm}
\caption{Online Cost Policy}
\begin{algorithmic}
\Require Budget $B$, \# queries $Q$, Bin size $S$, LLM set $L$
\Ensure LLM selections for each query
\State $N \gets \lceil \frac{Q}{S} \rceil$ \ \ \ \ \ \ \textcolor{gray}{\% Number of bins} %\Comment{Number of bins}
\State $B_{\text{bin}} \gets \frac{B}{N}$ \ \ \ \ \ \ \textcolor{gray}{\% Budget per bin} %\Comment{Budget per bin}
\State $B_{\text{left}} \gets 0$ \ \ \ \ \ \ \textcolor{gray}{} %\Comment{Budget remaining}
\For{each bin $i$ in $1$ to $N$}
\State $z \gets 0$ \ \ \ \ \ \ \ \ \ \ \ \ \ \textcolor{gray}{\% Initialize budget utilization} %\Comment{Initialize budget utilization}
\State $B_{\text{left}} \gets B_{\text{left}} + B_{\text{bin}} $ \ \ \ \ \ \ \ \ \ \ \ \ \ \textcolor{gray}{\% Remaining budget} %\Comment{Remaining budget for current bin}
\For{each query $q_t$ in bin $i$}
\State $Q_{\text{left}} \gets$ number of remaining queries in bin
\State $E \gets \{l \in L : C^l \leq \frac{\cos(\hat{\psi}(q_t), \hat{\theta}^{t}_l)}{(\frac{UB.e}{LB})^{z} (\frac{LB}{e})}\}$ \textcolor{gray}{\% Fit LLMs} %\Comment{Eligible LLMs}
\If{$E$ is empty}
\State $B' \gets \frac{B_{\text{left}}}{Q_{\text{left}}}$ \ \ \ \textcolor{gray}{\% Adjusted query budget} %\Comment{Adjusted query budget}
\State $E \gets \{l \in L : C^l \leq B'\}$
\If{$E$ is empty}
\State \textbf{return} ``Insufficient budget"
\EndIf
\EndIf
\State $l^* \gets \arg \max_{l \in E} \cos(\hat{\psi}(q_t), \hat{\theta}^{t}_l)$ \ \ \ \textcolor{gray}{\% pick best LLM}
\State $B_{left} \gets B_{left} - C^{l}$ \ \ \textcolor{gray}{\% Update remaining budget}
\State $z \gets z + \frac{C^{l}}{B_{\text{bin}}}$ \ \ \textcolor{gray}{\% Update budget utilization}
\State \textbf{yield} $l^*$ \ \ \textcolor{gray}{\% Return selected LLM for current query}
\EndFor
\EndFor
\end{algorithmic}
\end{algorithm}
\vspace{-0.2cm}
\noindent In Section \ref{subsubsec:cost_policy} of the main paper, we introduced our online cost policy, enabling users to set a cost budget $B$ and distribute it across a defined number of queries $Q$. Building on that discussion, we present the corresponding algorithm in Algorithm \ref{algo:cost_policy} and provide a concise summary of its key steps below. The algorithm operates by dividing the total query budget $Q$ into $N$ bins of size $S$, where $N = \lceil\frac{Q}{S}\rceil$. Each bin is allocated a portion of the total budget $B$, denoted as $B_{\text{bin}} = \frac{B}{N}$. At the start of each bin, the algorithm adds $B_{\text{bin}}$ to the remaining budget $B_{\text{left}}$ (budget leftover from previous bins/timesteps). 
% For each query within a bin, the algorithm calculates the number of remaining queries and identifies eligible large language models (LLMs), $E$, from the set of all available LLMs $L$ based on their cost and their Upper Confidence Bound (UCB) values. 
For each query within a bin, the algorithm selects eligible large language models (LLMs), $E$, from the set $L$ whose costs $C^l \ (l \in L)$ are below a threshold. This threshold is sensitive to both the LLM $l$ and the current budget utilization $z$, and is given by $\frac{\cos(\hat{\psi}(q_t), \hat{\theta}^{t}_l)}{(\frac{UB.e}{LB})^{z} (\frac{LB}{e})}$. If no LLMs fit the allotted budget thresholds, the algorithm adjusts the budget per query to $\frac{B_{\text{left}}}{Q_{\text{left}}}$ ($Q_{\text{left}}$ is the number of queries remaining in the bin) and re-evaluates eligible LLMs. If still no LLMs are available, the algorithm terminates with an ``Insufficient budget" message. Otherwise, it selects the LLM with the highest expected reward (defined in Equation \ref{eqn:exp_reward}) in the set of eligible LLMs $E$, updates the remaining budget, and yields the selected LLM for the current query. This process repeats for each query in the bin, ensuring the budget is optimally utilized across the entire query set.

\section{Additional Experimental Results and Analysis}\label{sec:appx_experiments}

\subsection{Comparison with Supervised Binary LLM Router}\label{subsec:appx_comparison_with_hybrid_llm}

\begin{figure}[!ht]
  \centering
  \begin{minipage}{0.48\textwidth}
    \centering
    \vspace{-0.8cm}
    \includegraphics[width=\linewidth]{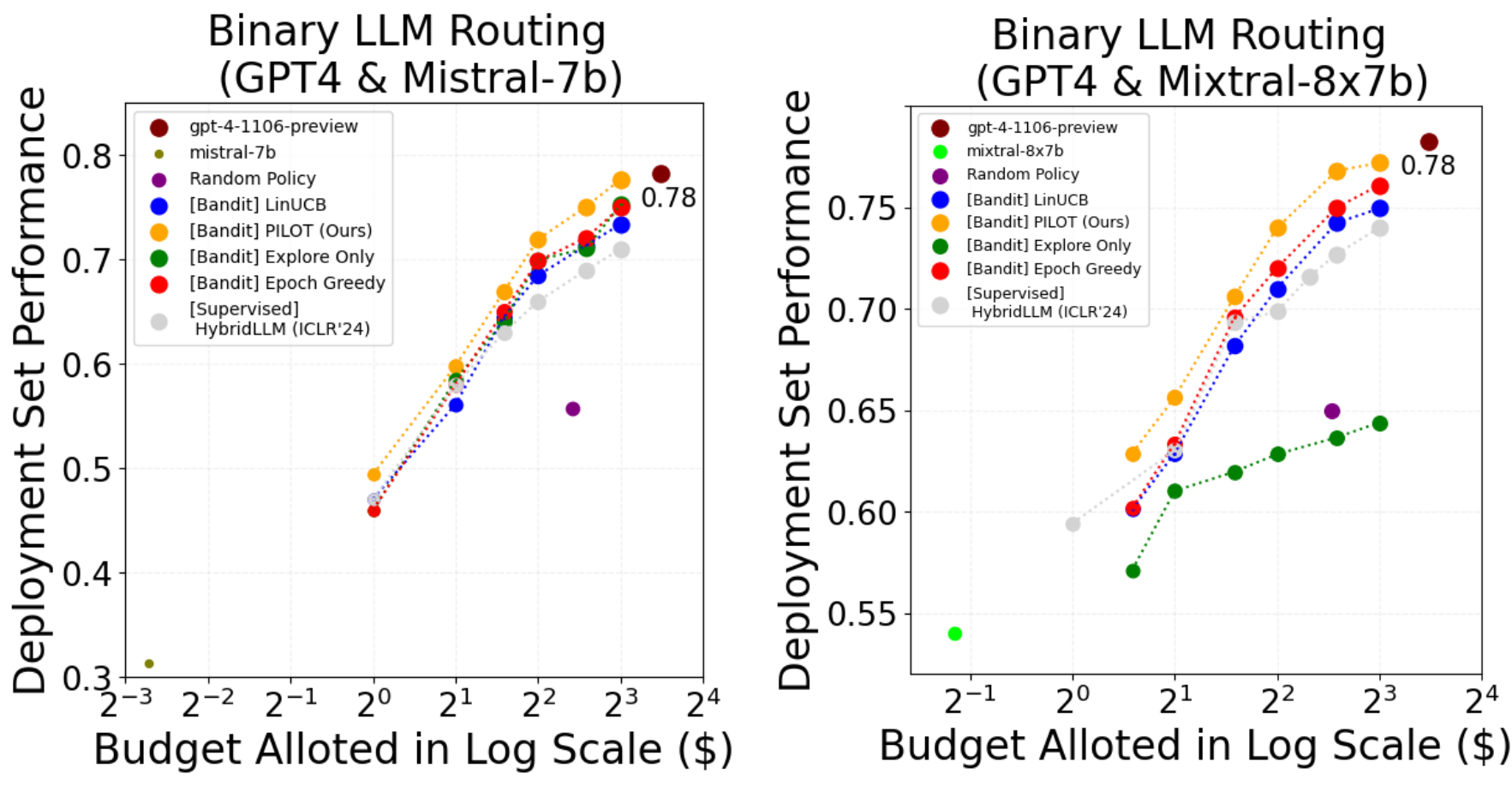}
        \vspace{-0.1cm}
        \caption{ \textbf{Performance vs Cost comparison with Supervised HybridLLM (Ding et al. 2024)} The left figure shows binary LLM routing comparisons for GPT-4 and Mistral-7b in the LLM pool. The right figure presents similar comparisons, this time for GPT-4 and Mixtral-8x7b. This study uses the Routerbench dataset}
      \label{fig:hybrid_LLM_comparison}
    % \vspace{-0.3cm}
  \end{minipage}
  \hspace{5.5cm}
  \begin{minipage}{0.48\textwidth}
    \centering
    \includegraphics[width=\linewidth]{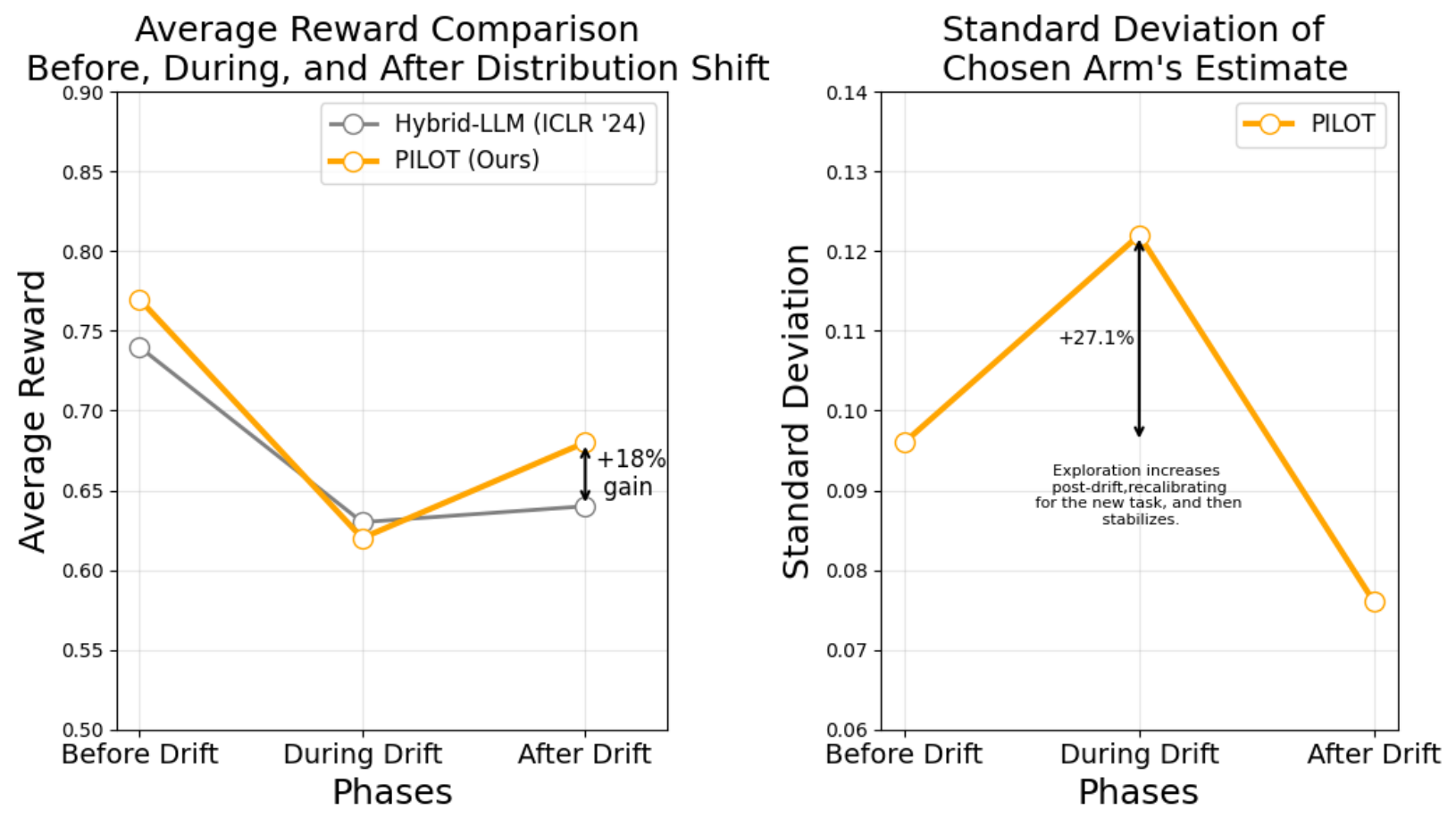}  
    \caption{ \textbf{Adaptability to Shift in Query Distribution} The left figure shows average reward comparison across different time instants - ``Before", ``During" and ``After" shift in query distribution from MMLU to GSM8k. The right figure shows that during the drift the exploration increases as new distribution is encountered and then the exploration settles down.}
    \label{fig:query_shift_adaptability}
  \end{minipage}
\end{figure}

\noindent As stated in Section \ref{subsubsec:baselines}, here (in Figure \ref{fig:hybrid_LLM_comparison}) we compare our bandit router \name \ with the state-of-the-art supervised router HybridLLM,  while keeping in mind that HybridLLM \cite{ding2024hybrid} relies on full supervision, whereas our approach operates with only bandit feedback arriving online. Furthermore, we quantitatively evaluate \name \ and HybridLLM, w.r.t. adaptability to shift in user queries to understand whether \name \ we improve upon existing routers in this aspect.
% \name owing to the online formulation of model routing, we study the impact of distribution shift on router performance.

\subsubsection{Performance vs Cost Analysis}
\label{subsec:appx_comparison_with_hybrid_llm_perf_vs_cost}
We compare \name \ with the deterministic variant of HybridLLM \cite{ding2024hybrid}, which assumes that LLMs are deterministic functions mapping input features to a single point in the output space. We do not experiment with the probabilistic variants due to the high cost involved—they require $20$ times more LLM calls to train the router, making them expensive to implement. These probabilistic variants utilize soft labels for training rather than hard labels, which necessitates sampling $10$ responses from each LLM (thus $20$ calls for two LLMs) per query and calculating a sample average. 
% Consequently, this process demands $20$ times the number of LLM calls ($10$ calls per LLM per query).

\noindent We use the Routerbench dataset for this comparison and pick two LLM combinations (GPT4 \& Mistral-7b, and, GPT4 \& Mixtral-7x8b) for binary routing task. We pick these LLM combinations to compare routers in the presence of large (GPT-4), medium (Mixtral-8x7b) and small (Mistral-7b) size language models. Furthermore, as per the protocol in Section 3.1 of the main paper, we use $1000$ samples for hyperparamter tuning and the remaining samples is split into ``learning" and ``deployment" buckets with $10:1$ ratio. 

\noindent As can be seen from Figure \ref{fig:hybrid_LLM_comparison}, \name \  performs on par with, and occasionally surpasses, HybridLLM. This underscores the effectiveness of bandit routers, achieving strong results without requiring full supervision.

\subsubsection{Adaptability to Shift in Query Distribution}
\label{subsec:appx_adapt_to_shift_in_query_dist}
Here in Figure \ref{fig:query_shift_adaptability}, we simulate a task shift (from MMLU to GSM8k) to create a streaming dataset, tracking average reward \& exploration (standard deviation of chosen arm's estimate). We then compare rewards before (``Before"), at (``During") and $5000$ steps after the transition (``After"). As can be seen in Figure \ref{fig:query_shift_adaptability} (left), \name \  adapts significantly post-drift, unlike the static supervised baseline. We also observe in Figure \ref{fig:query_shift_adaptability} (right) that the exploration of \name \ increases during drift and subdues after it, as expected.

\subsection{Query Complexity Analysis}
\label{appx:query_complexity_analysis}
In the context of binary LLM routing, an effective routing algorithm should allocate more complex queries to the more capable model (GPT-4) while routing simpler queries to the less expensive model (Mistral-7B/Mixtral-8x7B) \cite{ding2024hybrid}, to optimize performance within the given budget constraints. We thus investigate how our algorithm routes queries of varying complexity when faced with such a binary choice. For this analysis, we fix the overall budget for routing to \$$4$ and examine the average complexity of queries directed to each LLM. To quantify query complexity, we use \textit{Evol Complexity} \cite{liumakes} measure, which is useful for selecting hard samples for LLM alignment in comparison to scores such as LLM response \textit{perplexity} and \textit{Direct Scoring} \cite{chenalpagasus}. 

\begin{table}[!ht] % 'r' for right side and '0pt' to let LaTeX decide the width
\renewcommand{\arraystretch}{1.1} % Increase row height
\centering
\footnotesize
% \captionsetup{font=footnotesize}
\vspace{-0.2cm}
\hspace{-0.2cm}
\begin{tabular}{c@{\hskip 0.4cm}c@{\hskip 0.2cm}c>{\columncolor{blue!10}}c>{\columncolor{blue!10}}c}
\toprule
\textbf{LLM Pool} & \multicolumn{2}{c}{\textbf{LinUCB}} & \multicolumn{2}{>{\columncolor{blue!10}}c}{\textbf{\name}} \\
 & \textbf{QC} & \textbf{p-value} & \textbf{QC} & \textbf{p-value} \\
\midrule
GPT4 - Mistral7B & 2.61 & 0.06 & \textbf{2.64} & \textbf{0.004} \\
GPT4 - Mixtral8x7B & 2.62 & 0.05 & \textbf{2.75} & \textbf{5e-37} \\
\bottomrule
\end{tabular}
\vspace{-0.1cm}
\caption{\textbf{Query Complexity Analysis for Routed Queries:} \textit{QC} refers to average complexity of GPT-4 routed queries, \&, \textit{p-value} is from a Mann-Whitney U Test on average query complexity scores obtained from LLMs in the pool.}
\label{tab:query_complexity_analysis}
\vspace{-0.25cm}
\end{table}

% We find that the average complexity of queries routed to GPT-4 ($\mu_{GPT-4} = X$) is significantly higher than those routed to Mistral-7B ($\mu_{Mistral-7B} = Y$), with a p-value of $Z$ in a Mann-Whitney U test. 

% Furthermore, we find \name ... % compare with LinUCB to add more impact.

\noindent The average query complexity of GPT-4 routed queries (\textit{QC}) in Table \ref{tab:query_complexity_analysis} indicates that on average \name \
routes more complex queries to GPT-4, than LinUCB. Furthermore, the difference between the average query complexity of GPT-4 routed queries and Mistral-7B/Mixtral-8x7B routed queries is statistically significant for \name \ (Mann-Whitney U test's p-value $< 0.05$), unlike LinUCB. This indicates that \name \ not only considers the budget constraints but also assesses query complexity to make informed routing decisions.

% \vspace{-2mm}
\subsection{Analysis of Human Preference Learning}
\label{appx:human_pref_analysis}
Next, we analyze our offline human preference learning algorithm's (Section 2.2.1) accuracy of predicting the human preferred LLM for a given query. For this analysis, we use a subset of $500$ samples from the ChatArena \cite{chiang2024chatbot} dataset (related to the $11$ LLMs in Routerbench) that was not seen during training. We compare our accuracy with RouteLLM \cite{ong2024routellm} which also uses human preference data for learning how to route. We find our algorithm achieves a higher accuracy of \textbf{65.0}, in comparison to RouteLLM's  (uses Matrix factorization) accuracy of $63.6$. 
% (averaged over three runs).
% \vspace{-6mm}

\subsection{Binary LLM Routing Results}\label{subsec:appx_addn_binary_llm_routing}
\begin{figure}[ht]
  \centering
    \includegraphics[width=0.9\columnwidth]{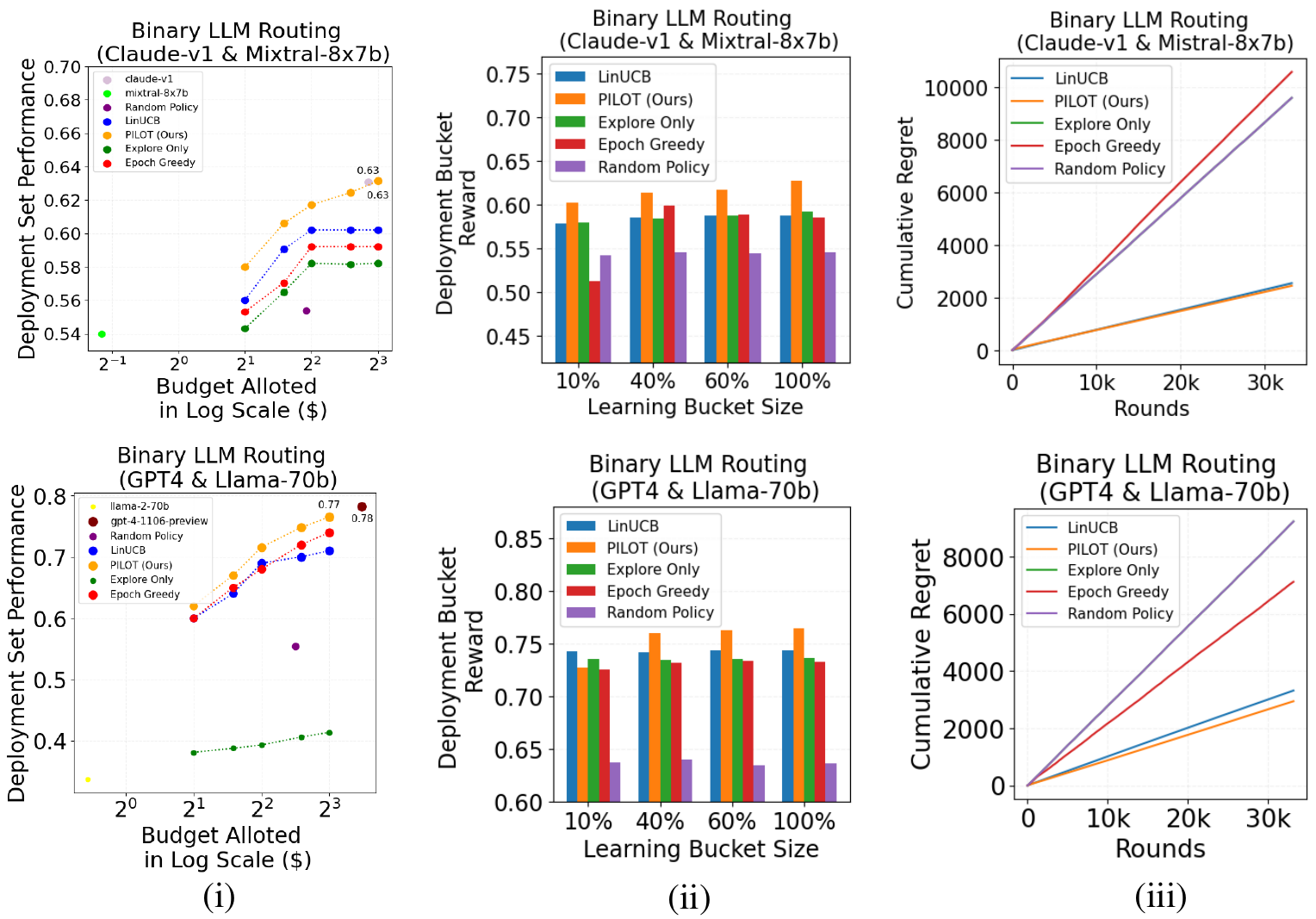}
    % \includesvg[width=1.0\columnwidth]{images/appendix/appendix_claude_v1_mixtral_gpt4_llama70b}
    \vspace{-0.1cm}
  \caption{\textbf{Bandit Feedback based Binary LLM Routing Evaluation}: Figure reports results for multi task data source setting (Routerbench). In the top row we have Claude-v1 and Mixtral-7x8b LLMs in the LLM pool, and in the botton row we have GPT-4 and Llama2-70b LLMs in the LLM pool. Column $(i)$ represents performance vs cost curves on the held-out deployment set; Column $(ii)$ represents performance across different learning bucket sizes; Column $(iii)$ represents cumulative regret. Our method \name \ , shown in \textcolor{orange}{orange}, performs the best across the metrics.}
  \label{fig:bin_routing}
  \vspace{-0.25cm}
\end{figure}

In Figure \ref{fig:single_domain_res} of the main paper, we reported results for binary routing with two LLM combinations - GPT4 \& Mistral-7b, and GPT4 \& Mixtral-7x8b. Here, in Figure \ref{fig:bin_routing}, we report results with two more LLM combinations (GPT4 \& Llama2-70b, Claude v1 \& Mixtral – 7x8B) to further evaluate our methods performance.  

\noindent Similar to our observation in the main paper, we find that across metrics, \name \ performs better than baselines.

% \begin{figure}[!ht]
%   \centering
%     \includegraphics[width=0.55\columnwidth]{images/appendix/query_shift_adaptability-cropped.pdf}
%     % \includesvg[width=1.0\columnwidth]{images/appendix/appendix_claude_v1_mixtral_gpt4_llama70b}
%   \caption{\textbf{Adaptability to Shift in Query Distribution}}
%   \label{fig:query_shift_adaptability}
%   %\vspace{-0.35cm}
% \end{figure}

\newpage
\subsection{Ablation and Sensitivity Analysis}\label{subsec:appx_addn_ablation_and_sens_anal}

% \section{Additional Binary LLM Routing Results}
\vspace{-0.2cm}
\begin{table}[ht]
    \centering
    \small
    \begin{minipage}{0.48\textwidth}
        \renewcommand{\arraystretch}{1}
        \centering
        % \footnotesize
        \vspace{-0.2cm}
        \hspace{-0.2cm}
        \begin{tabular}{>{}c@{\hskip 0.4cm}>{\columncolor{blue!3}}c@{\hskip 0.2cm}>{\columncolor{blue!6}}c >{\columncolor{blue!10}}c c}
            \toprule
            \textbf{Budget} & \textbf{\makecell{Pre-trained\\ Router}} & \textbf{\makecell{Pre-trained +\\10\% Online}} & \textbf{\name} & \textbf{LinUCB} \\
            \midrule
            \$1   & 0.34 & 0.61 & 0.63 & 0.60 \\
            \$1.5 & 0.34 & 0.61 & 0.66 & 0.63 \\
            \$2   & 0.35 & 0.63 & 0.69 & 0.64 \\
            \$3   & 0.38 & 0.65 & 0.73 & 0.68 \\
            \bottomrule
        \end{tabular}
        \vspace{-0.1cm}
        \caption{\textbf{Ablation Study:} Performance vs. Cost Comparison for pretrained router and effect of online training}
        \label{tab:ablation_performance_vs_cost}
    \end{minipage}
    % \vspace{-0.5cm}
    \hspace{0.52cm} % Added horizontal space
    \begin{minipage}{0.48\textwidth}
        \renewcommand{\arraystretch}{1.1}
        \centering
        % \footnotesize
        % \vspace{-0.9cm}
        % \hspace{-0.2cm}
        \begin{tabular}{c c c}
            \toprule
            \textbf{$\alpha$} & \textbf{\name} & \textbf{LinUCB} \\
            \midrule
            10 & 0.600 & 0.601 \\
            5  & 0.610 & 0.623 \\
            \rowcolor{blue!10} 2  & 0.645 & 0.649 \\
            1  & 0.641 & 0.640 \\
            \bottomrule
        \end{tabular}
        % \vspace{-0.1cm}
        % \caption{\textbf{Sensitivity Analysis of $\alpha$}}
        \caption{\centering \textbf{Sensitivity Analysis of $\alpha$}}
        \label{tab:sensitivity_alpha}
    \end{minipage}
% \vspace{-0.25cm}
\end{table}
% \vspace{-0.5cm}
%add a para to contextualize this analysis w.r.t. main paper
In Figure \ref{fig:single_domain_res} of the main paper, we analyzed the \name's performance of across cost budgets, and reward \& regret across rounds of online learning. Here we study the goodness of our pre-trained router (preference data based router) on the Routerbench deployment set. Please note, the preference data, Chatarena \cite{chiang2024chatbot}, is different from the Routerbench dataset \cite{hu2024routerbench}. Next, we also study the effect of exploration parameter (used in UCB computation in \name \ and LinUCB) on reward on the tuning dataset. 

\subsubsection{Ablation Analysis}
Here, in Table \ref{tab:ablation_performance_vs_cost} the initial pre-trained router shows strong foundational performance, and incorporating just 10\% online data enables it to adapt quickly and approach the performance of LinUCB, especially under constrained budgets. Notably, \name consistently outperforms the other approaches, even in low-cost scenarios.

\subsubsection{Sensitivity Analysis}
Table \ref{tab:sensitivity_alpha} presents the effect of the exploration parameter $\alpha$ on average reward, evaluated on the tuning dataset described in Section 3.1. Consistent with findings by \cite{li2010contextual}, the results exhibit an inverted U-shape. Small values of $\alpha$ result in insufficient exploration, limiting the algorithm’s ability to discover optimal LLM-query matches. In contrast, excessively high $\alpha$ values lead to over-exploration, missing opportunities to maximize reward.

\end{document}